# Structural State Translation: Condition Transfer between Civil Structures Using Domain-Generalization for Structural Health Monitoring

[1]Furkan Luleci, [1]F. Necati Catbas[*]

[1]Doctoral Student, Department of Civil, Environmental, and Construction Engineering, University of Central Florida, Orlando, Florida, 32816, USA (Email: furkanluleci@knights.ucf.edu)

[2*]Professor, Department of Civil, Environmental, and Construction Engineering, University of Central Florida, Orlando, Florida, 32816, USA (Corresponding author, email: catbas@ucf.edu)

**Abstract.** Using Structural Health Monitoring (SHM) systems with extensive sensing arrangements on every civil structure can be costly and impractical. Various concepts have been introduced to alleviate such difficulties, such as Population-based SHM (PBSHM). Nevertheless, the studies presented in the literature do not adequately address the challenge of accessing the information on different structural states (conditions) of dissimilar civil structures. The study herein introduces a novel framework named "*Structural State Translation*" (SST), which aims to estimate the response data of different civil structures based on the information obtained from a dissimilar structure. SST can be defined as *"Translating a state of one civil structure to another state after discovering and learning the domain-invariant representation in the source domains of a dissimilar civil structure"*. SST employs a Domain-Generalized Cycle-Generative (DGCG) model to learn the domain-invariant representation in the acceleration datasets obtained from a numeric bridge structure that is in two different structural conditions: *State-α* and *State-β*. After training DGCG on the source domains of *State-α* and *State-β* of a particular bridge, the model is used to generalize and transfer its knowledge (the "domain-invariant representation") to other structurally dissimilar bridges. In other words, the model is tested on three dissimilar numeric bridge models to translate their structural conditions. The evaluation results of SST via Mean Magnitude-Squared Coherence (MMSC) and modal identifiers showed that the translated bridge states (synthetic states) are significantly similar to the real ones. As such, the minimum and maximum average MMSC values of real and translated bridge states are 91.2% and 97.1%, the minimum and the maximum difference in natural frequencies are 5.71% and 0%, and the minimum and maximum Modal Assurance Criterion (MAC) values are 0.998 and 0.870. This study is critical for data scarcity and PBSHM, as it demonstrates that it is possible to obtain data from structures while the structure is actually in a different condition or state.

## 1. Introduction

Civil engineering structures age over time due to external effects such as environmental and man-made effects or stressors which negatively impact the safety of structures; particularly, the old structures are more vulnerable to such external effects. Consequently, a great deal of research and development is devoted to maintaining and improving the health conditions of civil structures and enabling effective structural asset sensing, control, and management strategies [1].

In practice, civil structure condition assessment techniques are challenging, as they are labor-intensive, subjective, time-consuming, and depend solely on human effort. Structural Health Monitoring (SHM) is suggested to complement current techniques [2,3] by tracking and assessing the condition of civil and others using data acquisition systems through various sensors such as cameras, lasers, and Non-Destructive Technique/Evaluation (NDT/E) [4–6]. SHM utilize sensorial data for damage diagnosis and prognosis for decision-making [7]. The decision-making phase is implemented to establish the structure's design validation, efficient maintenance, operations, and management strategies. This is achieved after the successful integration of SHM with sophisticated models within a structural identification framework [8].

### 1.1. Problem Definition

Although civil structures are worth monitoring for structural safety and operation, using SHM systems on every structure is economically not feasible. It is well-known that data collection procedures from civil structures can be challenging and costly, limiting the amount of information flow obtained from structures significantly. As a result, the data scarcity phenomenon is a critical issue in SHM due to the difficulties arising in data collection procedures. On the other hand, sensor- or transmission-based errors are also very typical during the monitoring process, resulting in information loss. The fact that SHM consists of data-driven applications makes the phenomenon of data scarcity even more significant. In that regard, utilizing information and estimating the structural health condition of civil structures based on the SHM-based information acquired from different structures will be an instrumental yet challenging SHM goal. In this regard, the problem addressed in this study can be illustrated in **Figure 1**. As structural systems can have quite complex mechanisms, they may have countless amounts of combinations of states (or conditions) which can include varying degrees of damage combined with different damage classes at various locations. Estimating these numerous amounts of combinations of structural states may not be intrinsically trivial and may not be addressed with currently available knowledge in the literature. Therefore, in the problem taken in this study, the structural states are simplified first into two different states (later in the study, developed into three states), *State-α* (healthy structure) and *State-β* (unhealthy structure – structural damage exists).

For instance, when the information (e.g., sensorial response data) is available for *State-α* while it is not for *State-β* of a *Structure A*, or similarly, when the information is available for *State- β* while it is not for State-α of a *Structure Z* (which is a different structure than *Structure* A), the objective would be to obtain those unavailable set of information of these structures. However, the studies presented in the literature do not adequately address this challenge: accessing the information (e.g., response data) of different structural states (conditions) of dissimilar structures. Nevertheless, several related works observed in the literature which use a population-based approach for the health monitoring of structures.

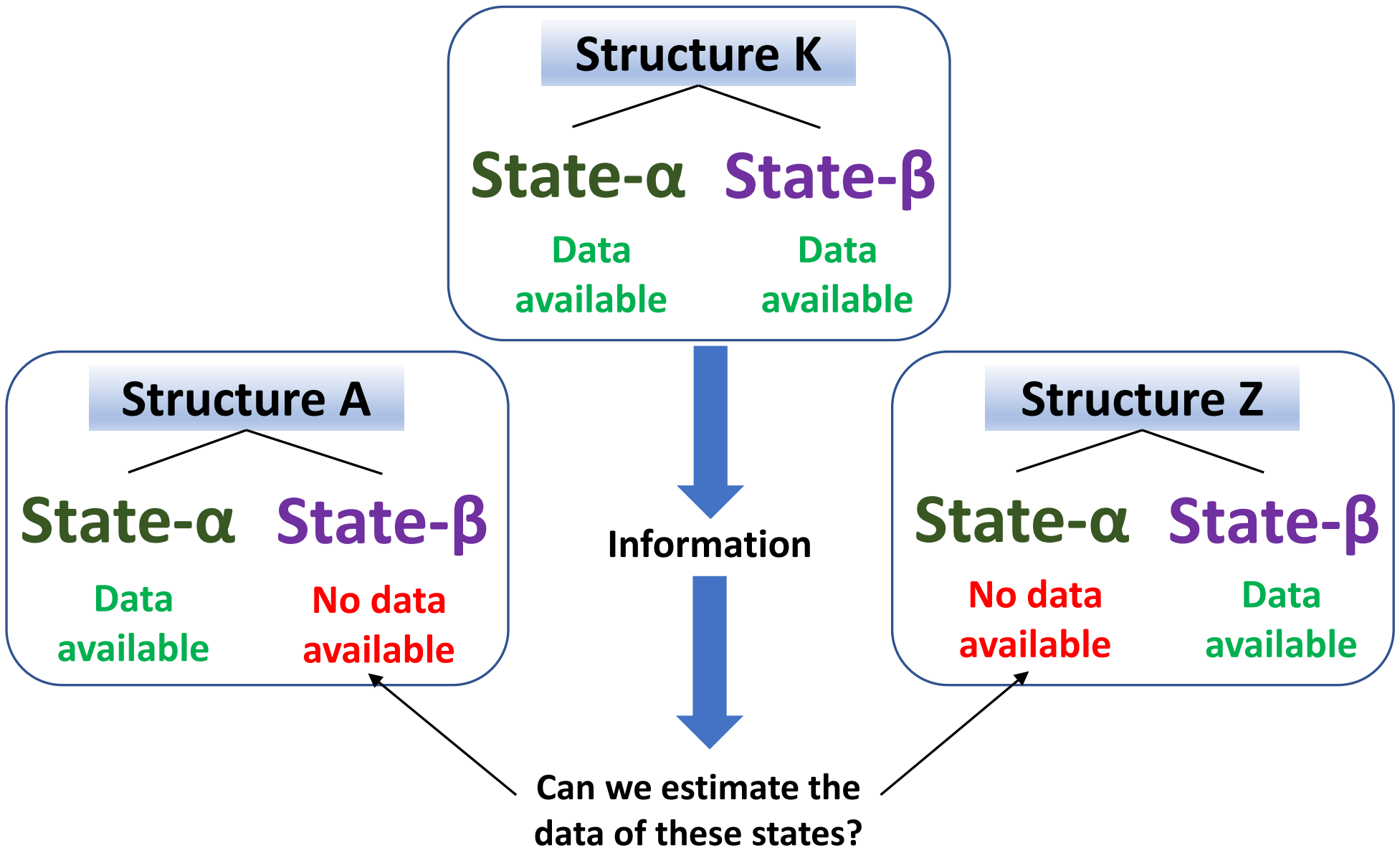

**Figure 1.** The research question addressed in this study: Can we estimate the data of different states of civil structures based on the knowledge from dissimilar structure(s)?

Condition assessment of populations of civil structures has been an area of interest for more than a decade. Some early studies evaluated nearly 2000 old bridges using SHM and modeling data [9]. More recently, the population-based concept has been introduced as PBSHM [10,11]. The objective of PBSHM is to increase the availability of physics-based and data-driven information on one set of civil structures based on the knowledge of other but "similar" populations of civil structures. Therefore, it is necessary to develop an abstract representation of each structure or population of structures based on similarity [12]. The term "similar" is an essential criterion for PBSHM, as it defines what knowledge is transferred to the other population for estimating the other populations' conditions. Two types of populations are suggested: homogeneous populations are defined as nominally-identical structures; heterogeneous populations [10,13,14] are defined as structures that contain disparate elements (while they may be similar, they are not identical). Knowing the similarity of the two structures requires comparing a structural representation of the structures; therefore, using graphs to define the structural similarity is found to be beneficial [15,16].

The transfer of information between structures happens in a feature representation space between different or similar domains [17]; hence, domain adaptation or similar techniques could be applied [18]. The information transferred for potential structural states from one structure to another contains different kinds of data representing different types and severities of damage at varying locations [19]. For that, having large amounts of data from various homogeneous and heterogenous populations of structures is vital, which could enable the use of feature representations from one set to another set [20]. In PBSHM, Transfer Learning (TL) and Domain Adaptation (DA) approaches are used for information transfer between structures. In one study [21], two heterogeneous TL approaches, kernelized Bayesian TL and heterogeneous feature augmentations, are compared and explored for PBSHM. In another study, DA is used to tackle the challenge of repair in SHM [22]. While the previous two studies tested on an aircraft wing, a study [23] proposed using heterogeneous TL with inconsistent feature spaces for PBSHM, testing on several populations of numerical and experimental civil structures. In another study, static alignment is proposed for DA and tested on heterogeneous populations of bridge structures [24]. More recently, another study has been conducted for transferring damage localization information between different aircraft wings using DA [25].

### 1.2. Motivation and Objective

For the last decade, the research and development in the civil SHM field have been progressive owing to the increasing use of Machine Learning (ML) methods as they have demonstrated practical solutions to various challenges faced in the SHM field [26–34]. In a standard statistical learning algorithm setting, a model is trained on the *i.i.d.* assumption, meaning training and testing data are identically and independently distributed. Empirically, the classic learning methods are optimized by minimizing the errors during the training, which greedily learn all the correlations identified for accurate prediction in inference. Although this is an effective solution in *i.i.d.* circumstances, this would deteriorate the learning algorithm's performance under distributional shifts, as shown in several studies [35–37]. As a result, the *i.i.d.* assumption is likely to be violated in real-life problems. For example, in civil SHM, a typical damage detection classifier trained on the data from a bridge pile will perform poorly on the hangers or when tested on a different structure because the classifier has not previously seen those data domains. Due to the covariate shift and/or semantic (label) shift between different domains, the classifier's performance is deteriorated when tested on a domain that is Out-Of-Distribution (OOD) [38].

While collecting data from all possible domains to train the learning algorithms is not feasible, different research activity areas are developed to deal with these shifts in target distributions, such as Multi-Task Learning (MTL), Transfer-Learning (TL), Domain-Adaptation (DA), Test-Time Learning (TTL), Zero-Shot Learning (ZSL), Domain-Generalization (DG), Meta-Learning and so on. MTL trains on several domains using a single model to make the model perform better on the same tasks that the model was trained on [39]. TL trains a model with a source domain and aims to transfer the knowledge of a trained model from one problem/domain/task to another related one by using a pretraining-finetuning strategy where the target samples are annotated [40]. DA has been broadly studied in the literature and is closest to DG. The main difference between DA and DG is that DA leverages additional data from test data, whether annotated or not and exposes related information about the target domain to the model [41–43]. TTL, also named test-time adaptation, is somewhere between DA and DG [44]. ZSL aims to learn the shifts in different test distributions; however, the shift is caused by label changes, while in DG, it is both label space changes and covariate shifts [45].

In summary, while TL, DA, and TTL have access to the test data in one form or another during the training, DG has no access to the training data, which makes it more challenging than the others, providing more realistic and favorable results in real-world applications [46]. DG aims to train a learner on single or many sources such that it generalizes to target data at testing time, where it is unavailable [47] which generally provides an upper bound for the performance of DG techniques [48]. Meta-Learning, on the other hand, is an emerging area where the aim is to learn the learning algorithm by learning from previous tasks' experience [49]. See **Table 1** for the general overview of the learning methods [50], which shows the number of source domains, whether the source and target domain distributions and labels are equal or not, and whether there is access to the training data during the training of the model.

**Table 1.** General overview of learning methods used to deal with distribution shifts [50].

| Learning Methods | $K$ | | $D_{XY}^{S}$ vs $D_{XY}^{T}$ | | $y_S$ vs $y_T$ | | Access to $D_X^T$ in training? |
|---|---|---|---|---|---|---|---|
| | =1 | >1 | $=$ | $\neq$ | $=$ | $\neq$ | |
| Multi-Task Learning | ✓ | ✓ | ✓ | | ✓ | | |
| Transfer Learning | ✓ | ✓ | | ✓ | | ✓ | ✓ |
| Zero-Shot Learning | ✓ | | | ✓ | | ✓ | |
| Domain Adaptation | ✓ | ✓ | | ✓ | ✓ | ✓ | ✓ |
| Test-Time Learning | ✓ | ✓ | | ✓ | ✓ | | ✓† |
| Domain Generalization | ✓ | ✓ | | ✓ | ✓ | ✓ | |

$K$: Number of source domains/tasks. $D_{XY}^{S/T}$: Source/target domain. $y_{S/T}$: Source/target label space. $D_X^T$: Target data.
†: Limited in quantities.

Considering the studies presented in the literature (explained in Section 1.1) using DA and TL, the overall objective is to adapt or transfer the knowledge from one set of the population of structure(s) to another set of population. The solutions offered in those studies for homogenous populations and, more recently, heterogeneous populations have shown some success with promising results. While PBSHM is a relatively new research area and a small portion of studies exist, there is a need for additional studies investigating to increase the availability of physics and/or data-driven knowledge on one set of civil structures based on the knowledge of another. In addition, while few studies are presented in the literature employing TL, DA, and ZSL learning approaches (as shown in Section 1.1), using MTL, TTL, and particularly DG is not observed.

DG is more feasible and practical when used in real-life scenarios than other learning methodologies, such as DA and TL, since DG does not use target data in one way or another, unlike DA and TL approaches. The studies in [51–54] theoretically demonstrated and improved that the feature representations that are invariant to different domains are general and transferable to different domains. Likewise, DG's objective is to decrease the discrepancy in representations between multiple source domains in a particular feature space to be domain-invariant. Thus, this enables the learned model to be generalizable to the other unseen domains. Motivated by the discussion presented above about the challenges in SHM and the trends in ML to deal with OOD, the work presented in this study aims to estimate the response data of different civil structures based on the information obtained from a dissimilar structure. On this basis, the objective of this study is established on the problem illustrated in **Figure 1**. The rest of the paper is outlined as follows: In the next section, the solution approach taken in this study is briefly explained, where Structural State Translation (SST) is presented, and the details of SST are described. Then, the dataset used in this study is explained, followed by a detailed description of the SST applied to the bridge models in this paper, such as the model used and the training procedure and so on. After that, an evaluation of the implementation results of SST is presented. Subsequently, the study is concluded with a summary of the work, results and conclusions.

### 1.3. Solution Approach

The study introduces a novel framework of "*Structural State Translation*" (SST). SST can be defined as *"*Translating a state of one civil structure to another state after discovering and learning the domain-invariant representation in the source domains of a different civil structure*"*. The framework employs a Domain-Generalized Cycle-Generative (DGCG) model to learn the domain-invariant representation in the source domains $D_{State-\alpha}^{Structure\ K}$ and $D_{State-\beta}^{Structure\ K}$ of *Structure K*, where the structure has two different structural states (conditions): *State-$\alpha$* and *State-$\beta$*. When the domain-invariant feature space is invariant to the different domains, the knowledge the model learned after the training is generalizable and transferable to other domains, as discussed in the previous section [51]. Taking advantage of this approach, the DGCG model that learned the domain-invariant representation in domains $D_{State-\alpha}^{Structure\ K}$ and $D_{State-\beta}^{Structure\ K}$, is used to generalize and transfer its knowledge to other domains. As such, DGCG is tested on the target domain (unseen) of *Structure A* where $D^{Structure\ A}_{State-\alpha}$ is translated (or generated) to $D^{Structure\ A}_{State-\hat{\beta}}$, where the hat "^" above $\beta$ denotes that it is the translated state. Similarly, the model is tested on the target domain (unseen) of *Structure Z* where $D^{Structure\ Z}_{State-\beta}$ is translated (or generated) to $D^{Structure\ Z}_{State-\hat{\alpha}}$ (**Figure 2)**. SST can also be carried out the other way around, depending on the availability of the states of structures, such as *Structure A* could only have *State-β* and *Structure Z* could only have *State-α* available or different combinations could be created. Intrinsically, more states and structures may be available in the source or the target domain. As a result, the model and/or training-related adjustments may be needed, which will be in the scope of future studies. A representative illustration of the SST process is shown in **Figure 2**.

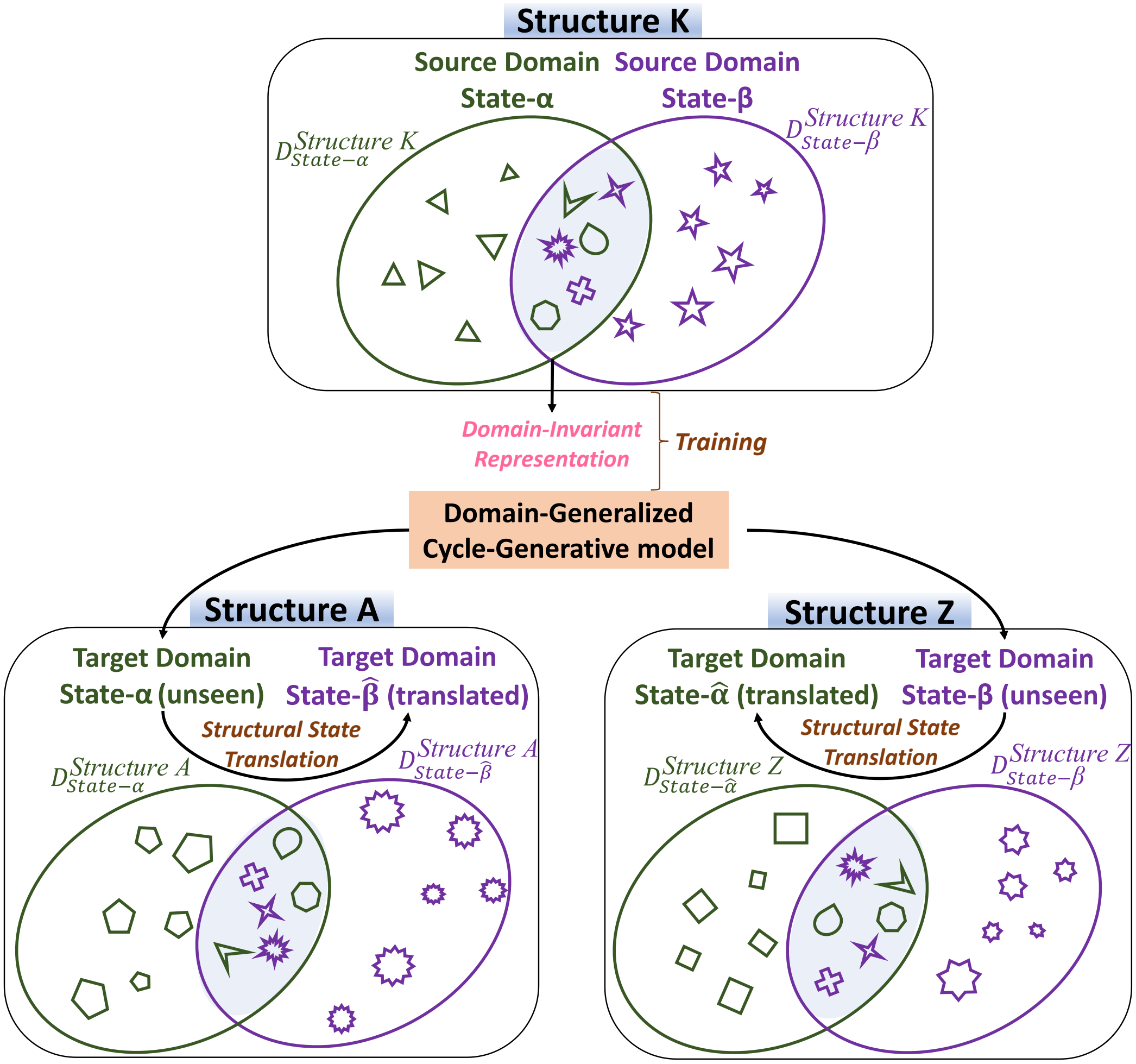

**Figure 2.** The solution approach taken in this study, "Structural State Translation", aims to learn the domain-invariant representation between different domains to generalize and transfer the learned knowledge to other structures whose domains are dissimilar to the source domain. The shapes illustrated in the ellipse diagrams are data samples belonging to its domain. While the domain of *State-α* has more deltoid-like samples, State-β consists of star-like samples. This is because when the structures are dissimilar to a "degree", the data domains of these structures show covariate and/or semantic shifts, causing OOD in their domains. Yet, when the domain-invariant feature space is invariant to different domains, that space is generalizable and transferable to other domains and remains invariant to any domain shifts.

**Side note.** Knowing the existence and the extent of domain-invariant representation in different domains remains an open question and active research area [55–59]. For example, the research by [56] argues that domain invariancy is inherited from both internal and mutual sides, where internal invariance exists in a single domain, and mutual invariance exists in multiple domains. Adapting this approach could be very beneficial for certain DG applications. Intuitively speaking, identical structures (homogeneous populations) or similar structures (heterogenous populations) should include domain invariancy; however, a domain relation, for example, between steel truss to prestressed concrete bridge structures or even bridge to dam structures, remains a big research question. Similarly, this statement is also valid for different damage types, severities, locations, or even the degree of linearities, as the structural nonlinearity in the structure may affect the data domains significantly.

The SST framework in this study is applied to four dissimilar numeric bridge models (which may belong to a heterogenous population). A numeric bridge model, *Bridge#1*, is adapted from an actual bridge structure and is used for training (source domain). The other three bridges, *Bridge#2*, *Bridge#3*, and *Bridge#4*, each modified significantly, are used for the test (unseen target domains). Each bridge dataset includes acceleration responses extracted from virtual sensor channels from the bridge models after applying the gaussian noise excitation signal to the models. A detailed explanation of the dataset used in this study is presented in the next section (Section 2). Nevertheless, briefly, the training dataset is created from a bridge model, *Bridge#1*, where it has two different structural states (conditions), *State-α* and *State-β*. *State-α* is the pristine condition, and *State-β* is the removal of the bottom steel chords from the sides close to the middle of the bridge (starting from the left - Section#11). A visual representation of the bridge states can be seen in **Figure 6**. As a result, *Bridge#1* yields two different but related domains $D_{State-\alpha}^{Bridge\#1}$ and $D_{State-\beta}^{Bridge\#1}$. Second, the test (or target) datasets, which are the datasets of *Bridge#2*, *Bridge#3*, and *Bridge#4*, are formed following the same approach as done for *Bridge#1* dataset.

Furthermore, another state is created, *State-γ*, where it is again the removal of the bottom steel chords from the sides close to the middle of the bridge but in a symmetrical position (starting from the left - Section#12) of *State-β*. The purpose of creating *State-γ* is to demonstrate that SST can be employed for geometrically symmetrical locations at the structures. As *State-β* and *State-γ* of the bridges are geometrically and materially symmetrical, it is expected that the translated *State-$\hat{\beta}$* from *State-α* and the translated *State-$\hat{\gamma}$* from *State-α* are the same in terms of their structural parameters. In other words, the translated *State-$\hat{\alpha}$* from *State- β* and the translated *State-$\hat{\alpha}$* from *State-γ* are expected to be the same. This phenomenon of symmetry is observed later in **Table 2** and **Table 3** as the frequencies, and the mode shapes of the bridges for *State-β* and *State-γ* are identical. As a result, *Bridge#2*, *Bridge#3*, and *Bridge#4* yield respectively domains $D^{Bridge\#2}_{State-\alpha}$, $D^{Bridge\#2}_{State-\beta}$, $D^{Bridge\#2}_{State-\gamma}$, and $D^{Bridge\#3}_{State-\alpha}$, $D^{Bridge\#3}_{State-\beta}$, $D^{Bridge\#3}_{State-\gamma}$, and $D^{Bridge\#4}_{State-\alpha}$, $D^{Bridge\#4}_{State-\beta}$, $D^{Bridge\#4}_{State-\gamma}$. Third, the DGCG model is trained in an unsupervised manner on the source domains $D^{Bridge\#1}_{State-\alpha}$ and $D^{Bridge\#1}_{State-\beta}$ to learn the domain invariant representation between *State-α* and *State-β*. Then, the model is tested on the target bridge models for different state translation "scenarios", respectively. As such, for *Bridge#2*, in Scenario I, the DGCG model is used to translate *State-α* to *State-$\hat{\beta}$* (which is the synthetic *State-β*); in Scenario II, DGCG is used to translate *State-α* to *State-$\hat{\gamma}$* (which is the synthetic *State-γ*); in Scenario III, DGCG is used to translate *State-β* to *State-$\hat{\alpha}$* (which is the synthetic *State-α*); in Scenario IV, DGCG is used to translate *State-γ* to *State-$\hat{\alpha}$*. This process is also carried out for other scenarios for *Bridge#3* and *Bridge#4*. The SST process implemented for the bridge models in this paper is illustrated in **Figure 3**. The bridge models and the dataset are explained in the next section.

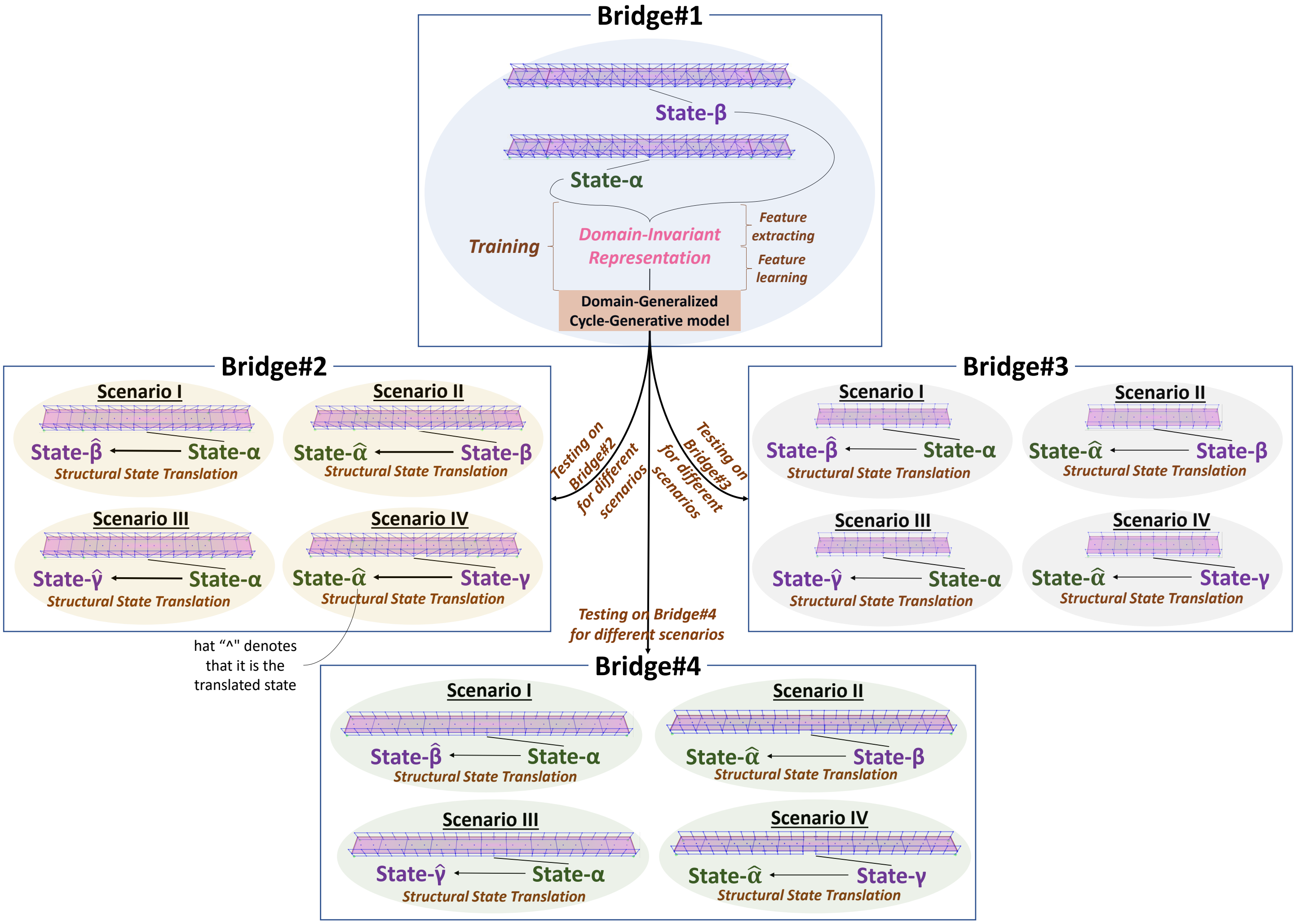

**Figure 3.** Structural State Translation applied to the numeric bridge models in this study. The state/condition of each bridge is translated to its other states via the Domain-Generalized Cycle-Generative model, which was trained with the datasets of *State-α* and *State-β* of *Bridge#1*.

## 2. Dataset

As mentioned above, the bridge models used in this study are numeric, modelled and analyzed in the Finite Element Analysis (FEA) program. Subsequently, the acceleration responses are extracted from each bridge model. After the bridges are modelled, the models are analyzed in two ways: Modal Analysis and Time History Analysis (THA). Modal Analysis is done for a general intuitive bridge similarity comparison between each model, and THA is carried out to extract the acceleration response signals from the bridge models. Subsequently, the acceleration response signals from the virtual sensor channels on the bridge models are extracted.

The *Bridge#1* model is adapted from a real steel truss footbridge structure located on the University of Central Florida campus. The footbridge comprises 177 ft long vertical truss frames connected in the middle span with a splice connection, spans 128 ft over a pond, and is 12 ft in width. The vertical truss members on the left and right sides are HSS10x10x3/8 for both top and bottom chords, and they are supported with HSS6x4x3/8 type vertical, HSS10x10x3/8 type vertical at the support and HSS4x4x1/4 type diagonal steel sections. Another truss system is used for lateral stability with HSS3x3x1/4 type diagonal cross braces and W12x22 type lateral beam elements. The bridge is separated into two spans, spliced at the middle with a plate connection, and carrying a 5 in thick layered aluminum-concrete composite deck. The bridge experiences light pedestrian traffic loads and small vehicles, e.g., golf carts. The structural drawings and the members used in the FEA program are given in **Figure 4**, where the elements are shown in the drawings with key letters.

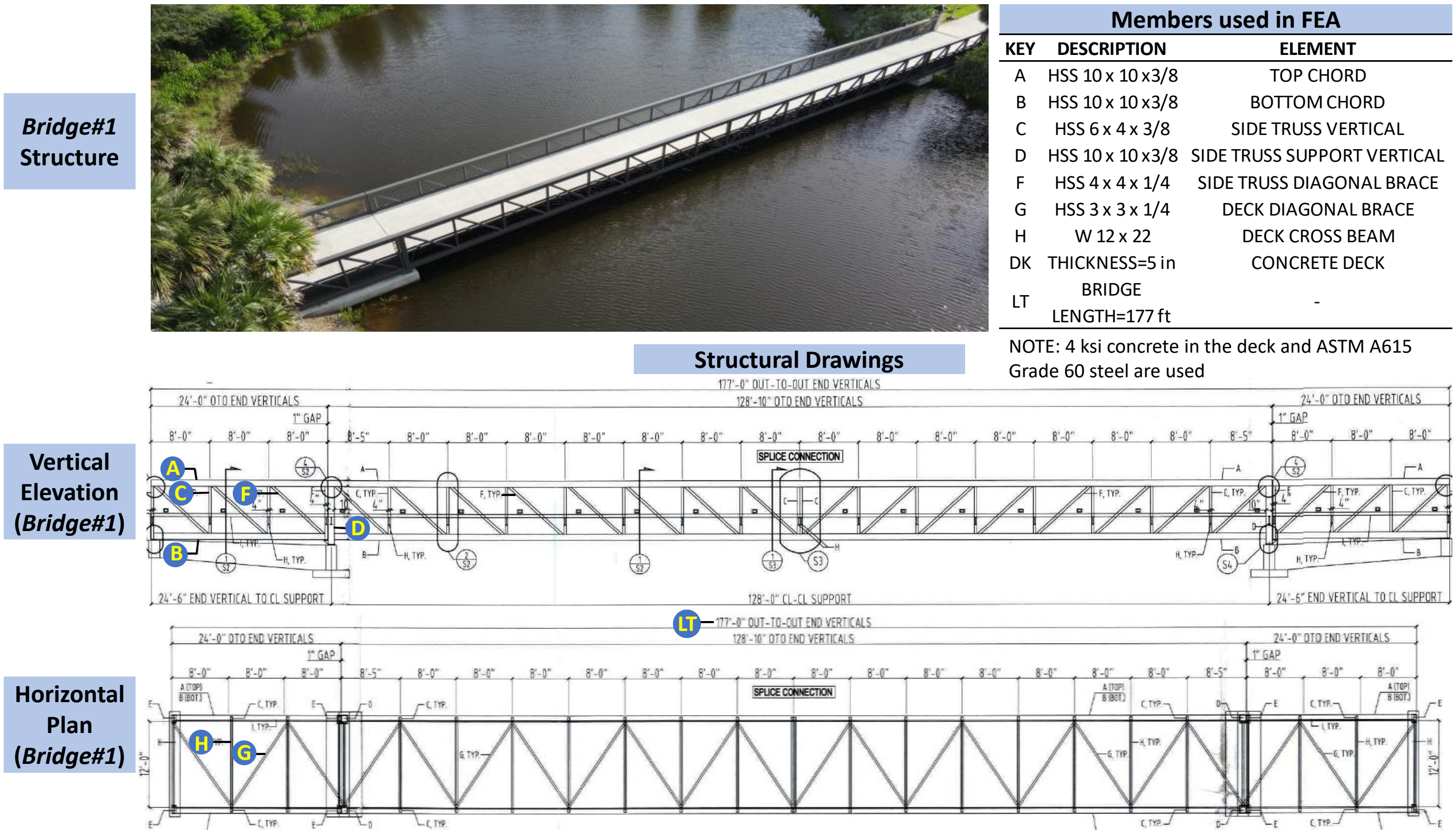

**Members used in FEA**

| KEY | DESCRIPTION | ELEMENT |
|---|---|---|
| A | HSS 10 x 10 x3/8 | TOP CHORD |
| B | HSS 10 x 10 x3/8 | BOTTOM CHORD |
| C | HSS 6 x 4 x 3/8 | SIDE TRUSS VERTICAL |
| D | HSS 10 x 10 x3/8 | SIDE TRUSS SUPPORT VERTICAL |
| F | HSS 4 x 4 x 1/4 | SIDE TRUSS DIAGONAL BRACE |
| G | HSS 3 x 3 x 1/4 | DECK DIAGONAL BRACE |
| H | W 12 x 22 | DECK CROSS BEAM |
| DK | THICKNESS=5 in | CONCRETE DECK |
| LT | BRIDGE LENGTH=177 ft | - |

NOTE: 4 ksi concrete in the deck and ASTM A615 Grade 60 steel are used

**Figure 4.** The structural drawings and members of the footbridge modelled as *Bridge#1* in the FEA program.

The other bridge models, *Bridge#2*, *Bridge#3*, and *Bridge#4*, on the other hand, were modelled in the FEA program with several structural adjustments. For instance, the removal of the deck-diagonal brace, reducing the thickness of the bridge deck to 2.5 in, and shortening the total bridge length by 48 ft and 10 in are the changes made in *Bridge#2*. In *Bridge#3*, the number of changes was increased: top and bottom chords were replaced with HSS5x5x.25; side truss diagonal and deck diagonal braces were removed; deck crossbeam was replaced with W10x15; the concrete deck thickness was increased to 6 in; lastly, total bridge length was reduced to 80 ft 2 in while fixed end parts were also removed. While keeping the changes made in key letters A, B, F, G, and DK (**Figure 5**) in Bridge#3, some modifications were made to H, DK, and LT in *Bridge#4*. As such, even-numbered deck crossbeams were removed (except the first and the last beams), and the remaining odd-numbered beams were replaced with W16x26; while keeping the bridge length the same as *Bridge#1*, and the interior supports were removed. As a result, the amount of modifications made to bridges goes from low to high as *Bridge#2*, *Bridge#3*, and *Bridge#4*. The modelled bridges can be seen in **Figure 5**.

First, modal analysis is performed using Ritz vectors as it provides a better participation factor, which is important for the speed of analysis. Then, the natural frequencies and mode shapes of each state of the bridge models are identified. New mode appearances/mode switches are shown with different colors when the state of the bridge is changed from *State-$\alpha$* to *State-$\beta$* or *State-$\gamma$*. **Table 2-3** show the mode shapes and natural frequencies identified for each state of the bridge models. Also, the bridge models are sorted according to their overall stiffness (based on *State-$\alpha$*). Generally, it is observed that *Bridge#2* is the stiffest and *Bridge#4* is the most flexible. On the other hand, the stiffness/flexibility of Bridge#1 and Bridge#3 are roughly similar, as shown in **Figure 7**. After identifying the modal parameters for each bridge model, the THA is performed. For THA, a time history function is defined as an excitation signal to apply to the bridge models in the FEA program, which is a Gaussian noise with mean $\mu = 0$ and standard deviation $\sigma = 0.3$. The excitation signal is applied to the bridge models for 1024 seconds (t), and its sampled frequency (fs) is 256 Hz, as shown in **Figure 7**. Correspondingly, the acceleration response signals were collected for $t = 1024$ seconds and $fs = 256$ Hz from the virtual sensor channels on each bridge model, as illustrated in **Figure 6**.

As seen in **Table 2-3**, the natural frequencies of the bridges are significantly different. The type of mode shapes, however, show more similarity to each other while having considerable differences in some other modes. Additionally, some differences in mode shapes (new mode appearance/mode switch) can be observed when the bridge's state is changed from *State-$\alpha$* to *State-$\beta$* or *State-$\gamma$*. On the other hand, there is no new mode appearance/mode switch in *Bridge#2* after the state changes from *State-$\alpha$* to *State-$\beta$* or *State-$\gamma$,* given that the bridge itself is the stiffest among other bridges. The Modal Analysis results indicate that the bridges are structurally and topologically dissimilar. Though some degree of similarity between bridges should exist, this remains an open question. A degree of similarity of the bridge structures in metric space could reveal an intuition about the similarity between the bridge models, which is an active research area in PBSHM, as mentioned previously [16].

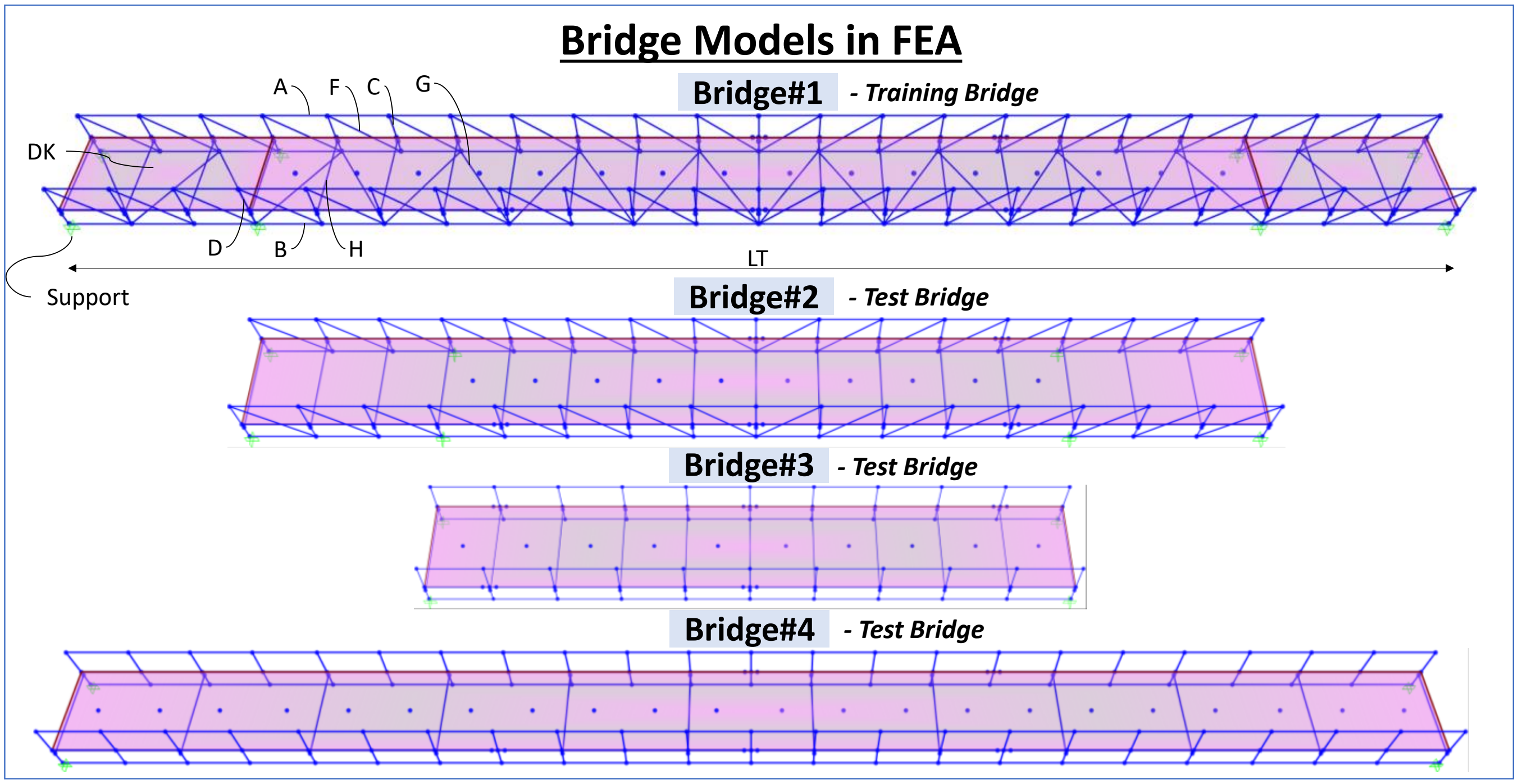

## Glossary

**Members used in FEA for Bridge#1**

| KEY | DESCRIPTION | ELEMENT |
|---|---|---|
| A | HSS 10 x 10 x3/8 | TOP CHORD |
| B | HSS 10 x 10 x3/8 | BOTTOM CHORD |
| C | HSS 6 x 4 x 3/8 | SIDE TRUSS VERTICAL |
| D | HSS 10 x 10 x3/8 | SIDE TRUSS SUPPORT VERTICAL |
| F | HSS 4 x 4 x 1/4 | SIDE TRUSS DIAGONAL BRACE |
| G | HSS 3 x 3 x 1/4 | DECK DIAGONAL BRACE |
| H | W 12 x 22 | DECK CROSS BEAM |
| DK | THICKNESS=5 in | CONCRETE DECK |
| LT | BRIDGE LENGTH=177 ft | - |

**Members used in FEA for Bridge#2**

| KEY | DESCRIPTION | ELEMENT | MODIFIED ELEMENT |
|---|---|---|---|
| A | HSS 10 x 10 x3/8 | TOP CHORD | - |
| B | HSS 10 x 10 x3/8 | BOTTOM CHORD | - |
| C | HSS 6 x 4 x 3/8 | SIDE TRUSS VERTICAL | - |
| D | HSS 10 x 10 x3/8 | SIDE TRUSS SUPPORT VERTICAL | - |
| F | HSS 4 x 4 x 1/4 | SIDE TRUSS DIAGONAL BRACE | - |
| G | HSS 3 x 3 x 1/4 | DECK DIAGONAL BRACE | Removed |
| H | W 12 x 22 | DECK CROSS BEAM | - |
| DK | THICKNESS=5 in | CONCRETE DECK | Thickness=2.5in |
| LT | TOTAL BRIDGE LENGTH=177 ft | - | Bridge Length=128 ft 2 in |

**Members used in FEA for Bridge#3**

| KEY | DESCRIPTION | ELEMENT | MODIFIED ELEMENT |
|---|---|---|---|
| A | HSS 10 x 10 x3/8 | TOP CHORD | Replaced with HSS 5 x 5 x .25 |
| B | HSS 10 x 10 x3/8 | BOTTOM CHORD | Replaced with HSS 5 x 5 x .25 |
| C | HSS 6 x 4 x 3/8 | SIDE TRUSS VERTICAL | - |
| D | HSS 10 x 10 x3/8 | SIDE TRUSS SUPPORT VERTICAL | - |
| F | HSS 4 x 4 x 1/4 | SIDE TRUSS DIAGONAL BRACE | Removed |
| G | HSS 3 x 3 x 1/4 | DECK DIAGONAL BRACE | Removed |
| H | W 12 x 22 | DECK CROSS BEAM | Replaced with W10 x 15 |
| DK | THICKNESS=5 in | CONCRETE DECK | Thickness=6 in |
| LT | TOTAL BRIDGE LENGTH=177 ft | - | Total Bridge Length=80 ft 2 in (fixed end parts are removed) |

**Members used in FEA for Bridge#4**

| KEY | DESCRIPTION | ELEMENT | MODIFIED ELEMENT |
|---|---|---|---|
| A | HSS 10 x 10 x3/8 | TOP CHORD | Replaced with HSS 5 x 5 x .25 |
| B | HSS 10 x 10 x3/8 | BOTTOM CHORD | Replaced with HSS 5 x 5 x .25 |
| C | HSS 6 x 4 x 3/8 | SIDE TRUSS VERTICAL | - |
| D | HSS 10 x 10 x3/8 | SIDE TRUSS SUPPORT VERTICAL | - |
| F | HSS 4 x 4 x 1/4 | SIDE TRUSS DIAGONAL BRACE | Removed |
| G | HSS 3 x 3 x 1/4 | DECK DIAGONAL BRACE | Removed |
| H | W 12 x 22 | DECK CROSS BEAM | Even numbered beams are removed and remaining beams are replaced with W 16 x 26 |
| DK | THICKNESS=5 in | CONCRETE DECK | Thickness=6 in |
| LT | TOTAL BRIDGE LENGTH=177 ft | - | Total Bridge Length=177 ft (interior supports are removed) |

**Figure 5.** The bridge models modelled in the FEA program and modifications are made to each bridge model, *Bridge#2*, *Bridge#3*, and *Bridge#4*, based on *Bridge#1* model.

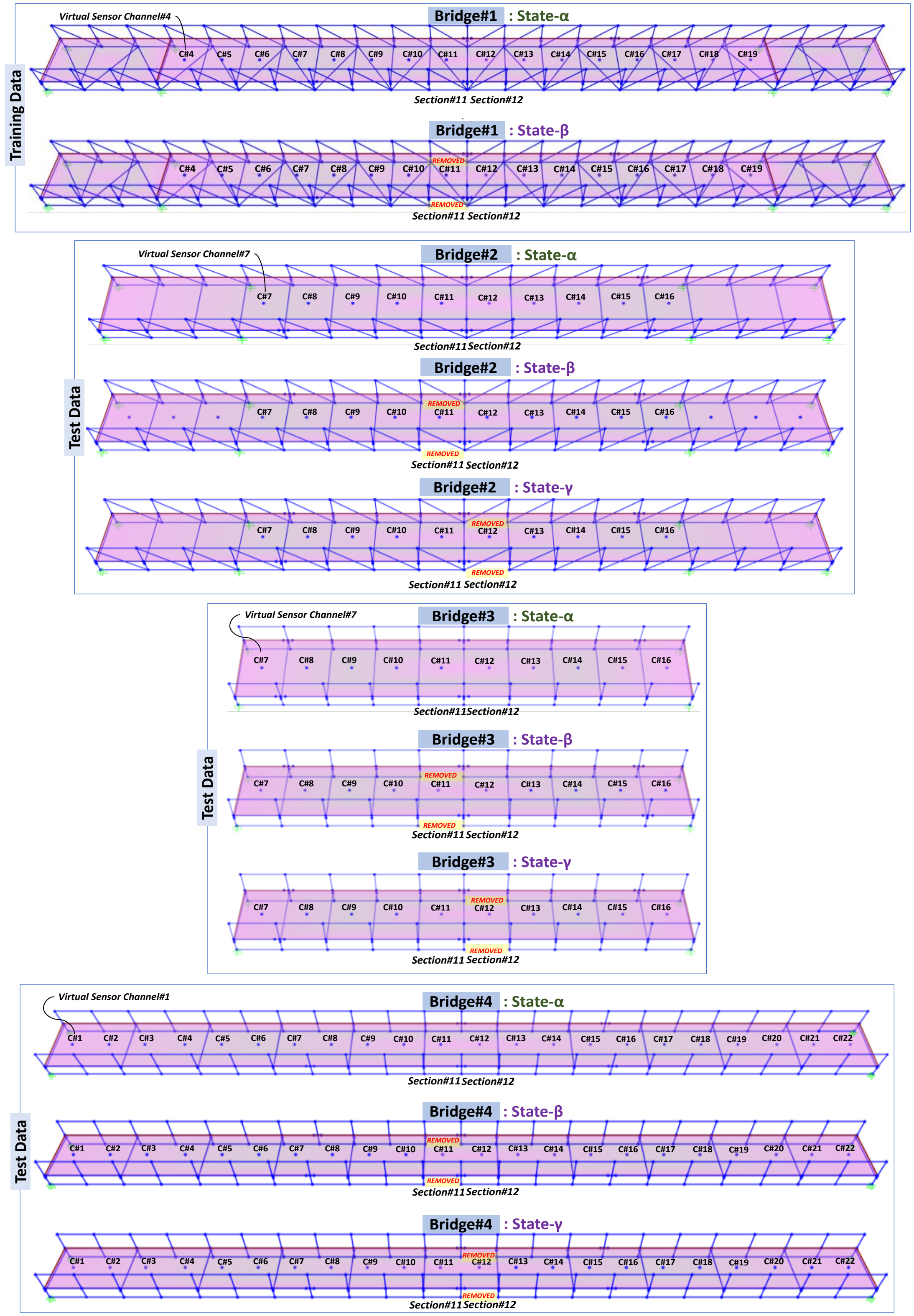

**Figure 6.** The bridge models and their states (pristine state: *State-α*, and removal of bottom chords from Section#11 and Section#12: *State-β* and *State-γ*). After the Time History Analysis, the acceleration response signals are extracted from each virtual sensor channel (denoted as "C"), and then the dataset of acceleration response signals for each bridge state is formed.

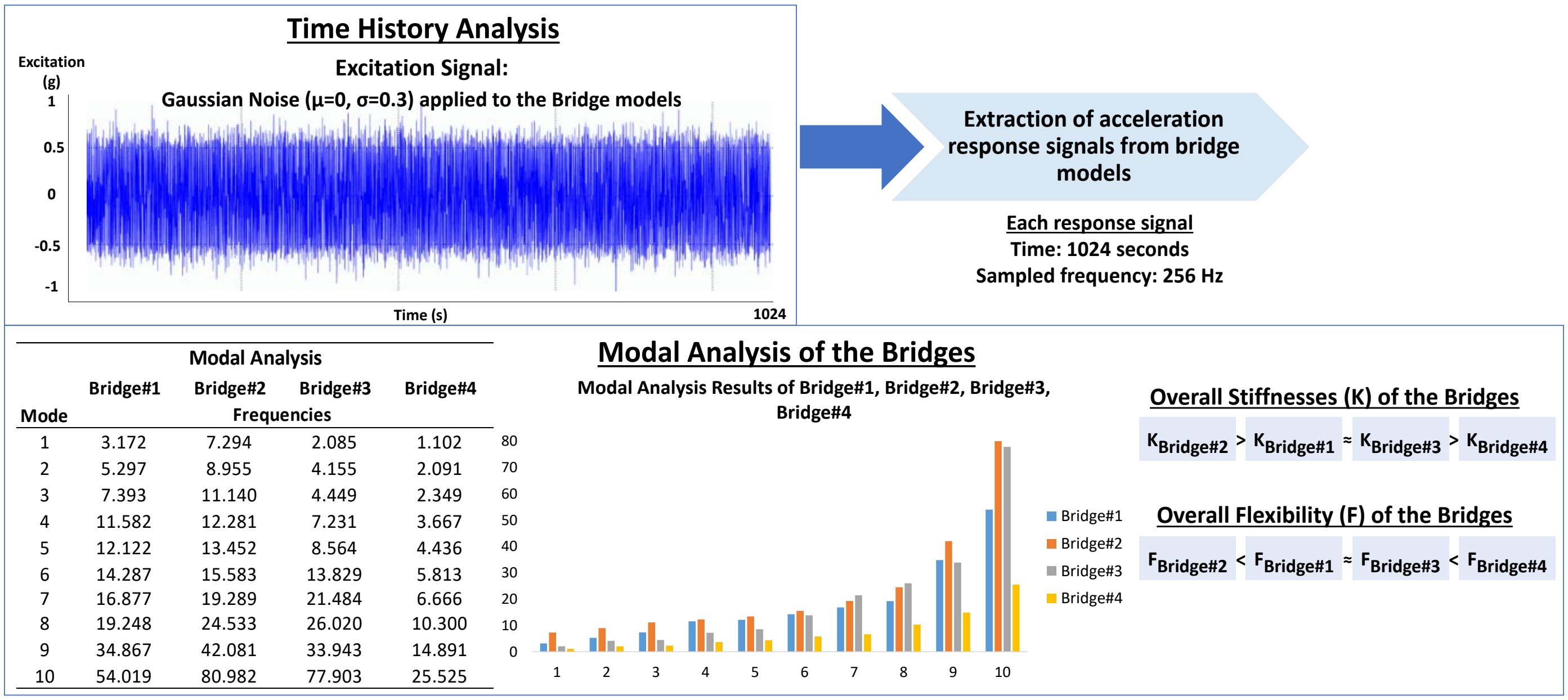

**Figure 7.** The Gaussian excitation signal applied to the bridge models for Time History Analysis. The natural frequency values after the modal analysis of the bridges and their stiffness/flexibility are shown in the figure.

**Table 2.** Natural frequencies and mode shapes of each state of *Bridge#1* and *Bridge#2*. New mode appearances/mode switches are shown with different colors when the state of the bridge is changed from *State-α* to *State-β* or *State-γ*.

**Bridge#1**

| Mode No | State-α f (Hz) | State-α Mode Shape | State-β f (Hz) | State-β Mode Shape | State-γ f (Hz) | State-γ Mode Shape |
|---|---|---|---|---|---|---|
| 1 | 3.17 | 1st bending | 2.89 | 1st bending | 2.89 | 1st bending |
| 2 | 5.30 | 1st torsional | 5.31 | 1st torsional | 5.31 | 1st torsional |
| 3 | 7.39 | 2nd bending | 7.26 | 2nd bending | 7.26 | 2nd bending |
| 4 | 11.58 | 3rd bending | 10.83 | 1st lateral & torsional | 10.83 | 1st lateral & torsional |
| 5 | 12.12 | 1st lateral | 11.15 | 2nd torsional | 11.15 | 2nd torsional |
| 6 | 14.29 | 4th bending | 13.92 | 4th bending | 13.92 | 4th bending |
| 7 | 16.88 | 5th bending | 16.15 | 5th bending | 16.15 | 5th bending |
| 8 | 19.25 | 6th bending | 19.04 | 6th bending | 19.04 | 6th bending |
| 9 | 34.87 | 7th bending | 31.73 | 7th bending | 31.74 | 7th bending |
| 10 | 54.02 | 8th bending | 52.51 | 8th bending | 52.52 | 8th bending |

**Bridge#2**

| Mode No | State-α f (Hz) | State-α Mode Shape | State-β f (Hz) | State-β Mode Shape | State-γ f (Hz) | State-γ Mode Shape |
|---|---|---|---|---|---|---|
| 1 | 7.29 | 1st bending | 6.90 | 1st bending | 6.90 | 1st bending |
| 2 | 8.96 | 1st torsional | 8.61 | 1st torsional | 8.61 | 1st torsional |
| 3 | 11.14 | 2nd bending | 11.14 | 2nd bending | 11.14 | 2nd bending |
| 4 | 12.28 | 3rd bending | 12.20 | 3rd bending | 12.19 | 3rd bending |
| 5 | 13.45 | 4th bending | 13.40 | 4th bending | 13.38 | 4th bending |
| 6 | 15.58 | 5th bending | 14.59 | 5th bending | 14.45 | 5th bending |
| 7 | 19.29 | 6th bending | 17.71 | 6th bending | 17.60 | 6th bending |
| 8 | 24.53 | 7th bending | 21.73 | 7th bending | 21.63 | 7th bending |
| 9 | 42.08 | 8th bending | 41.41 | 8th bending | 41.40 | 8th bending |
| 10 | 80.98 | 9th bending | 74.83 | 9th bending | 74.67 | 9th bending |

**Table 3.** Natural frequencies and mode shapes of each state of *Bridge#3* and *Bridge#4*. New mode appearances/mode switches are shown with different colors when the state of the bridge is changed from *State-α* to *State-β* or *State-γ*.

| | Mode No | State-α f (Hz) | State-α Mode Shape | State-β f (Hz) | State-β Mode Shape | State-γ f (Hz) | State-γ Mode Shape |
|---|---|---|---|---|---|---|---|
| Bridge#3 | 1 | 2.09 | 1st bending | 2.03 | 1st bending | 2.03 | 1st bending |
| | 2 | 4.15 | 1st torsional | 4.12 | 1st torsional | 4.12 | 1st torsional |
| | 3 | 4.45 | 2nd bending | 4.35 | 2nd bending | 4.35 | 2nd bending |
| | 4 | 7.23 | 3rd bending | 7.07 | 3rd bending | 7.07 | 3rd bending |
| | 5 | 8.56 | 2nd torsional | 10.18 | 4th bending | 10.18 | 4th bending |
| | 6 | 13.83 | 4th bending | 13.69 | 1st bending & longitudinal | 13.69 | 1st bending & longitudinal |
| | 7 | 21.48 | 5th bending | 13.97 | 1st longitudinal | 13.98 | 1st longitudinal |
| | 8 | 26.02 | 6th bending | 21.46 | 5th bending | 21.46 | 5th bending |
| | 9 | 33.94 | 7th bending | 31.20 | 6th bending | 31.21 | 6th bending |
| | 10 | 77.90 | 8th bending | 65.06 | 7th bending | 65.11 | 7th bending |

| | Mode No | State-α f (Hz) | State-α Mode Shape | State-β f (Hz) | State-β Mode Shape | State-γ f (Hz) | State-γ Mode Shape |
|---|---|---|---|---|---|---|---|
| Bridge#4 | 1 | 1.10 | 1st bending | 1.06 | 1st bending | 1.06 | 1st bending |
| | 2 | 2.09 | 1st torsional | 2.08 | 1st torsional | 2.08 | 1st torsional |
| | 3 | 2.35 | 2nd bending | 2.33 | 2nd bending | 2.33 | 2nd bending |
| | 4 | 3.67 | 3rd bending | 3.52 | 3rd bending | 3.52 | 3rd bending |
| | 5 | 4.44 | 2nd torsional | 4.51 | 2nd torsional | 4.51 | 2nd torsional |
| | 6 | 5.81 | 3rd torsional | 5.13 | 1st bending & torsional | 5.13 | 1st bending & torsional |
| | 7 | 6.67 | 4th bending | 6.57 | 4th bending | 6.57 | 4th bending |
| | 8 | 10.30 | 5th bending | 10.14 | 2nd bending & torsional | 10.14 | 2nd bending & torsional |
| | 9 | 14.89 | 6th bending | 14.36 | 3rd bending & torsional | 14.36 | 3rd bending & torsional |
| | 10 | 25.52 | 7th bending | 24.47 | 4th bending & torsional | 24.47 | 4th bending & torsional |

## 3. Structural State Translation

In this section, the SST framework applied in this paper is presented. The framework consists of four major phases: (1) Preprocessing, (2) Training, (3) Translation, and (4) Postprocessing. In the Preprocessing phase, the acceleration response signals extracted from the sensor channels in each bridge model and for each corresponding state are divided into 16-second tensors. In the Training phase, the DGCG model is trained on the source domains of *State-α* and *State-β* of *Bridge#1*. In the Translation phase, the DGCG model is used to translate the target domains (unseen) of *State-α*, *State-β*, *and State-γ* of each *Bridge#2*, *Bridge#3*, and *Bridge#4* to other corresponding domains of *State-*$\hat{\alpha}$, *State-*$\hat{\beta}$, and *State-*$\hat{\gamma}$, respectively. In the Postprocessing phase, the generated (or translated - synthetic) 16-second tensors in the domains of *State-*$\hat{\alpha}$, *State-*$\hat{\beta}$, and *State-*$\hat{\gamma}$ of each bridge are concatenated back to form the whole 1024 seconds signals in their corresponding domains of each state. The overview of the methodology is illustrated in **Figure 8**. In the following subsections, the phases are explained in detail.

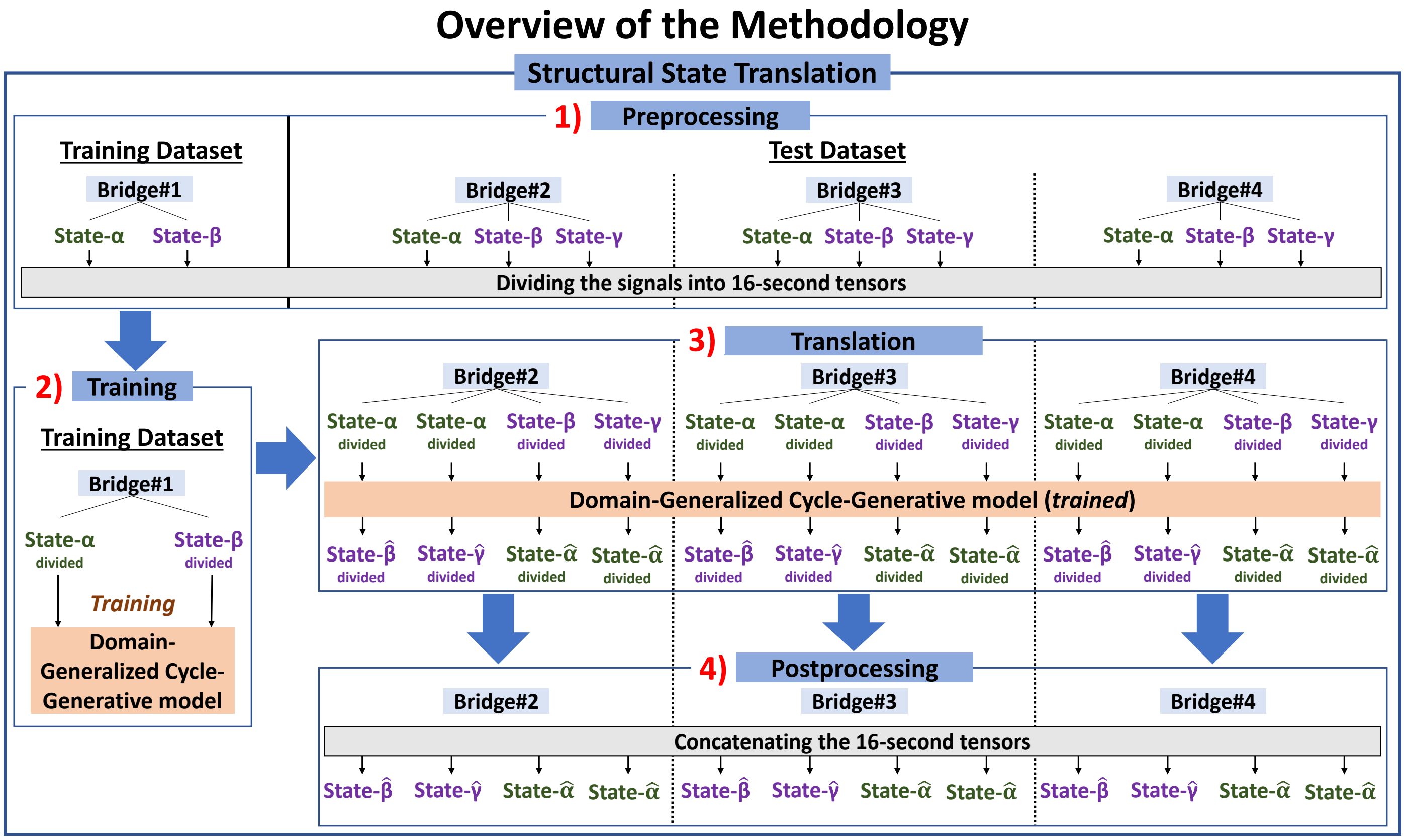

**Figure 8.** Overview of the methodology: The signals in the domains of each state of each bridge are first divided into 16-second tensors. Then, the Domain-Generalized Cycle-Generative model is trained on the domains of states of *Bridge#1*. Then, the model is tested on the domains of states of other bridges (unseen target domains), *Bridge#2*, *Bridge#3*, and *Bridge#4*, to translate their domains of each state to other bridge states. Finally, the generated 16-second tensors in the translated (synthetic) domains of each state are concatenated back into their corresponding domains of states.

### 3.1. Preprocessing

In the Preprocessing phase, the 1024-second signals extracted from the virtual sensor channels in each bridge model for each different state are divided into 16-second tensors, yielding 64 amount of 16-second tensors per sensor channel (**Figure 9**). For instance, *State-α* and *State-β* of *Bridge#1* each have 19 acceleration response signals from sensor channels, divided into 16-second tensors, resulting in *State-α divided* and *State-β divided*. The same procedure is also applied to *Bridge#2*, *Bridge#3*, and *Bridge#4*. The motive for dividing the signals is to increase the model's training efficiency. Feeding the whole signal to the DGCG model would not be feasible due to the large memory size requirement for processing the data samples during the training. The common practice in the Deep Learning field is using a smaller batch size during the model training due to the computational efficiency in training. For instance, feeding each 19 acceleration response signals extracted from *Bridge#1* in the DGCG model can be computationally very expensive and is not feasible since each signal has 262,144 data points ($1024\ (seconds) \times 256\ (Hz) = 262{,}144\ (data\ points)$). Therefore, a data division approach is implemented. The same division strategy is also successfully implemented in some studies, such as: [33,60,61]. Lastly, contrary to popular practice, normalization before feeding the data samples in the model is not implemented. Intuitively, feeding raw data into the model can conserve the features in the raw data and enhance the generative skill of the DGCG model [33,60,61]. But normalization would be needed if the data consists of large spikes.

Note that since the excitation signal applied to the bridge models is a constant Gaussian noise, the response signals extracted from the bridges are periodic, meaning the signal is repetitive at some intervals. This phenomenon is observed and verified in the frequency domains of 16-second tensors, as the 16-second tensors appeared to be very similar to each other after their division from the 1024-second signal.

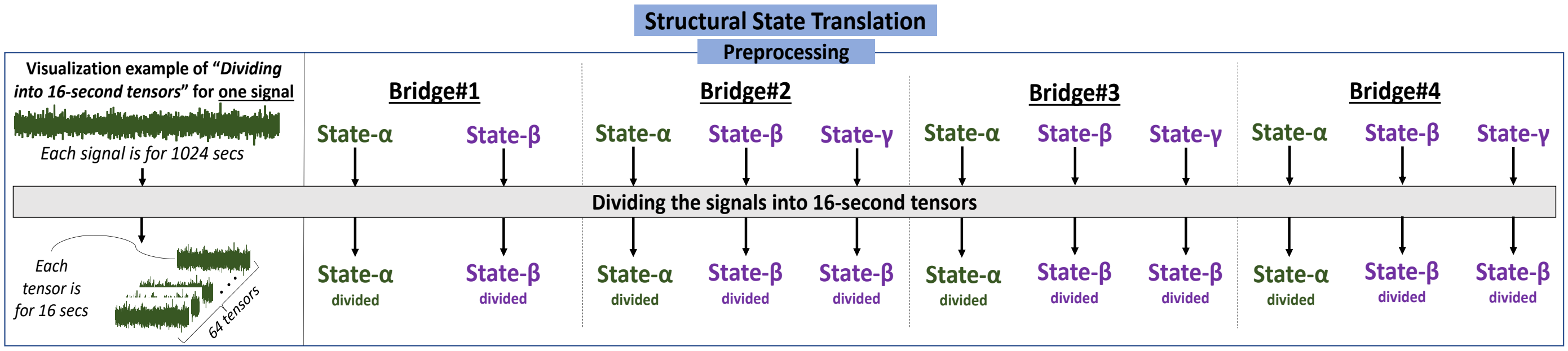

**Figure 9.** Preprocessing: 1024-second signals extracted from the virtual sensor channels in each bridge model for each different state are divided into 16-second tensors, yielding 64 16-second tensors.

## 3.2. Training

### 3.2.1. Domain-Generalized Cycle-Generative Model

The architecture of the Domain-Generalized Cycle-Generative (DGCG) model used (**Figure 10**) in this study consists of two generator and two critic networks (critic is discriminator in the vanilla Generative Adversarial Networks – GAN [62]), where each generator and critic consist of mapping networks. In essence, the model is built on the CycleGAN model [63], where two GAN networks train iteratively in a cycle-consistent fashion. Training of the DGCG model is visualized in **Figure 11**. The DGCG model in this study uses Wasserstein distance with gradient penalty in the objective function of critic, which was introduced as an improvement to the training of vanilla GAN [64,65] (the discriminator is named "critic" in their study). For instance, one study [66] conducted a comparative analysis that reveals the success of implementing Wasserstein distance in CycleGAN (CycleWGAN) and additionally using gradient penalty (CycleWGAN-GP), compared to regular CycleGAN.

The mapping networks (as shown in **Figure 10**) consist of a Gated Linear Unit (GLU) (Dauphin et al. 2016), which helps to learn the broader features via linear operation in the data instead of nearby features (convolution operation). When working on image-based applications, convolution operations are more advantageous as they inform the model about the vicinity of the interested pixel. However, using linear transformation operations is more beneficial for learning the signal-based data structures since it provides information about the broader features of the data [27]. Furthermore, skip connections (with adding) are employed in both mapping networks and generators, which were observed to be extremely helpful during the training [68]. Mish activation [69] function is used in both generator and critic, which was also utilized in YOLOv4 [70]. However, in GLUs used in the mapping network, sigmoid is employed as the activation function. At the end of the generator, the Tanh function is used. It is observed that using mapping networks, including GLUs, skip connections, and Mish activation functions, helped the DGCG model significantly during the training, where they assisted in diminishing the vanishing gradient problem, which is a significant challenge in the DL field. As a result, information that could be learned from the data during the training of DGCG is maximized. The rest of the details, such as filter, stride, and padding sizes used in the model, can be seen in **Figure 10**. Lastly, the parameters used in the model are given in **Table 4**. Note that 4096 is the number of data points in each 16-second tensor $(16\ (seconds) \times 256\ (Hz) = 4096\ (data\ points))$.

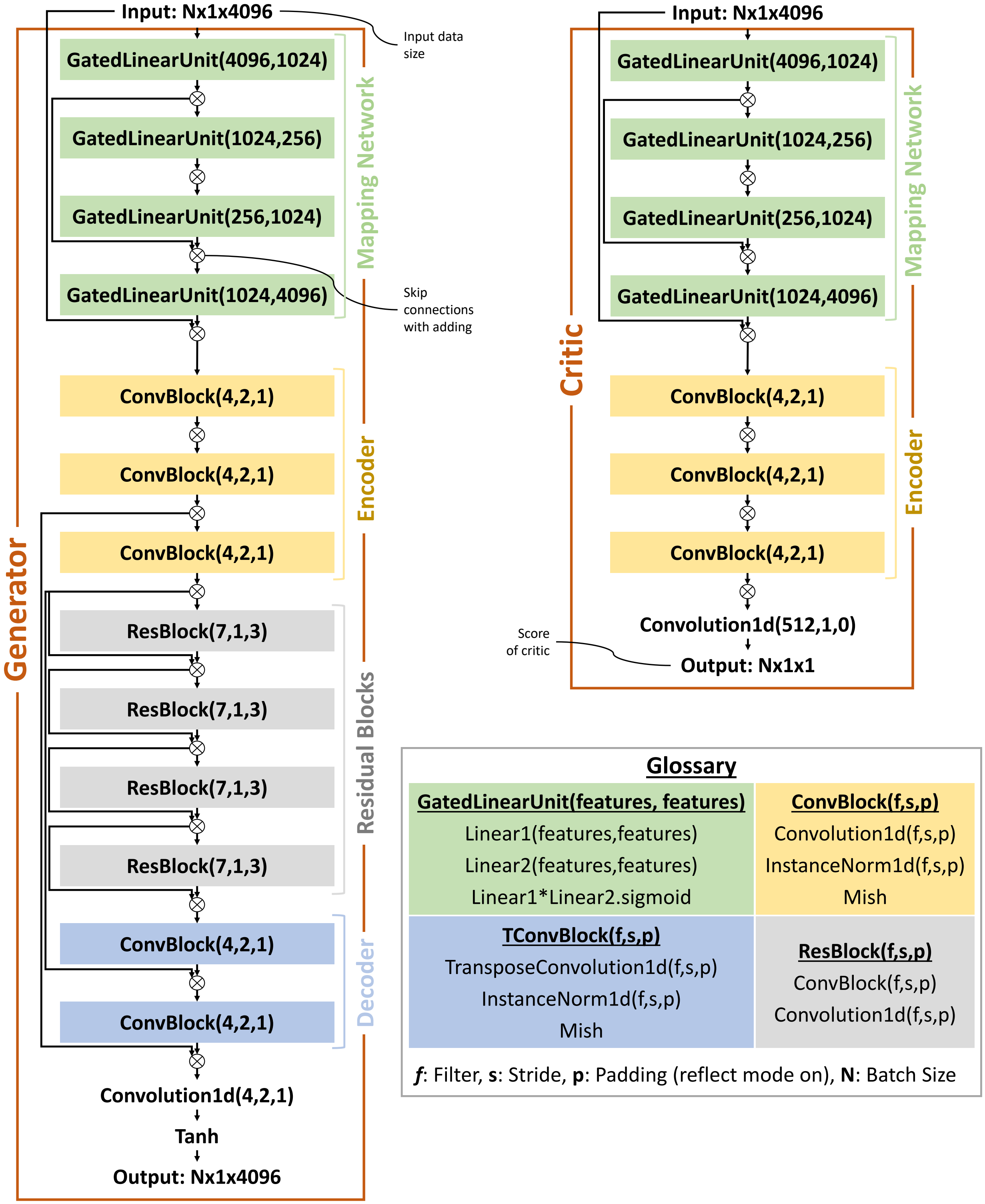

**Figure 10.** The single model architecture of the Domain-Generalized Cycle-Generative (DGCG) model: generator and critic, each including mapping networks consisting of gated linear units.

**Table 4.** Parameters used in the Domain-Generalized Cycle-Generative model

| Symbol | Description |
|---|---|
| Batch Size (N) | 4 |
| Epoch | 160 |
| Learning Rate | 1x10$^{-5}$ |
| Critic Iteration per Epoch | 10 |
| $\lambda_{Id}$ (Identity Loss) | 10 |
| $\lambda_{Id}$ (Identity Loss) | 10 |
| $\lambda_{Cyc}$ (Cycle Loss) | 10 |
| $\lambda_{GP}$ (Gradient Penalty) | 30 |
| Optimizer | AdamW |

#### 3.2.2. Training Data Workflow

As the authors defined in [71,72], domain alignment is one of the most common approaches used in the Domain Generalization (DG) research area, where the central idea is to minimize discrepancies among source domains to learn domain-invariant representations. The motivation is simple: features invariant to a covariate/semantic shift in the source domain should also be invariant to shift in any unseen target domain. Some techniques used for domain alignment are minimizing maximum mean discrepancy, minimizing KL divergence, minimizing contrastive loss, and domain-adversarial learning, which is one of the most popular research techniques [50]. As mentioned previously, the DGCG model used in this study is built based on CycleGAN, using Wasserstein distance and gradient penalty, to learn the domain invariant representation between two source domains: $D_{State-\alpha}^{Bridge\#1}$ and $D_{State-\beta}^{Bridge\#1}$. Thus, it can be said that this study belongs to the domain-adversarial learning for domain alignment in the DG area.

As mentioned, the aim is to have DGCG learn the domain-invariant representation in domains $D_{State-\alpha}^{Bridge\#1}$ and $D_{State-\beta}^{Bridge\#1}$. The learning is carried out in a cycle-consistent manner, where the model is iteratively trained on multi-domains to decrease the discrepancy in the representations between domains in a particular feature space to be domain-invariant. This enables the learned model to be generalizable and to transfer its knowledge to the other unseen domains. The DGCG model's training data workflow is shown in **Figure 11**, which can be followed along with the equations presented below. Each loss function used during the training is presented in Eq. (1) to Eq. (6) and can be followed in **Figure 11**. For the *State-$\alpha$* to *State-$\beta$* training flow, Eq. (1) is enforced to train the critic $C_{\alpha\beta}$, where $C_{\alpha\beta}$ receives 16-second tensor *x* from $D^{Bridge\#1}_{State-\beta}$ and 16-second synthetic tensor *x'* generated by $G_{\alpha\beta}$. The Wasserstein distance of the outputs of $C_{\alpha\beta}$ are computed with a gradient penalty function, where the mean squared distance via the L2 norm of the multiplication of the respective gradients of *x'* and the output from $C_{\alpha\beta}$ on *x'* are calculated. Eq. (3) is enforced for the training of the generator $G_{\alpha\beta}$, which is based on the minimization of the output received from the corresponding critic $C_{\alpha\beta}$. Eq. (5) enforces the cycle-consistency between two different state domains. For instance, when 16-second tensor *x* from $D^{Bridge\#1}_{State-\alpha}$ is translated by $G_{\alpha\beta}$ to 16-second tensor *x'*, the translated tensor *x'* should be the same as the original tensor after it is translated back by $G_{\beta\alpha}$ to 16-second tensor *x*, which is belonging to $D^{Bridge\#1}_{State-\alpha}$. Similarly, but the other way around, when 16-second tensor *x* from $D^{Bridge\#1}_{State-\beta}$ is translated by $G_{\beta\alpha}$ to 16-second tensor *x'*, the translated tensor *x'* should be the same as the original tensor after it is translated back by $G_{\alpha\beta}$ to 16-second tensor *x*, which is belonging to $D^{Bridge\#1}_{State-\beta}$. Eq. (6) is used to enforce the identity loss function. This function defines that when $G_{\alpha\beta}$ receives a 16-second tensor *x* from the domain $D^{Bridge\#1}_{State-\beta}$, it should output the same tensor *x* since $G_{\alpha\beta}$ already knows how to generate tensors belonging to $D^{Bridge\#1}_{State-\beta}$. The same statement is true the other way around for the *State-β* to *State-α* process. Note that in Eq. (6), L1 distance is used. Additionally, the procedure executed in Eq. (1) and Eq. (3) for *State-α* to *State-β* is also implemented the other way around by enforcing Eq. (2) and Eq. (4) to achieve *State-β* to *State-α*. The lambda parameters ($\lambda$) used in the equations Eq. (1), Eq. (2) and Eq. (8), for optional weight adjustments are given in **Table 4**. Finally, the DGCG model is trained iteratively based on the minimization of total critic losses Eq. (7) and total generator losses Eq. (8) using the AdamW optimizer [73].

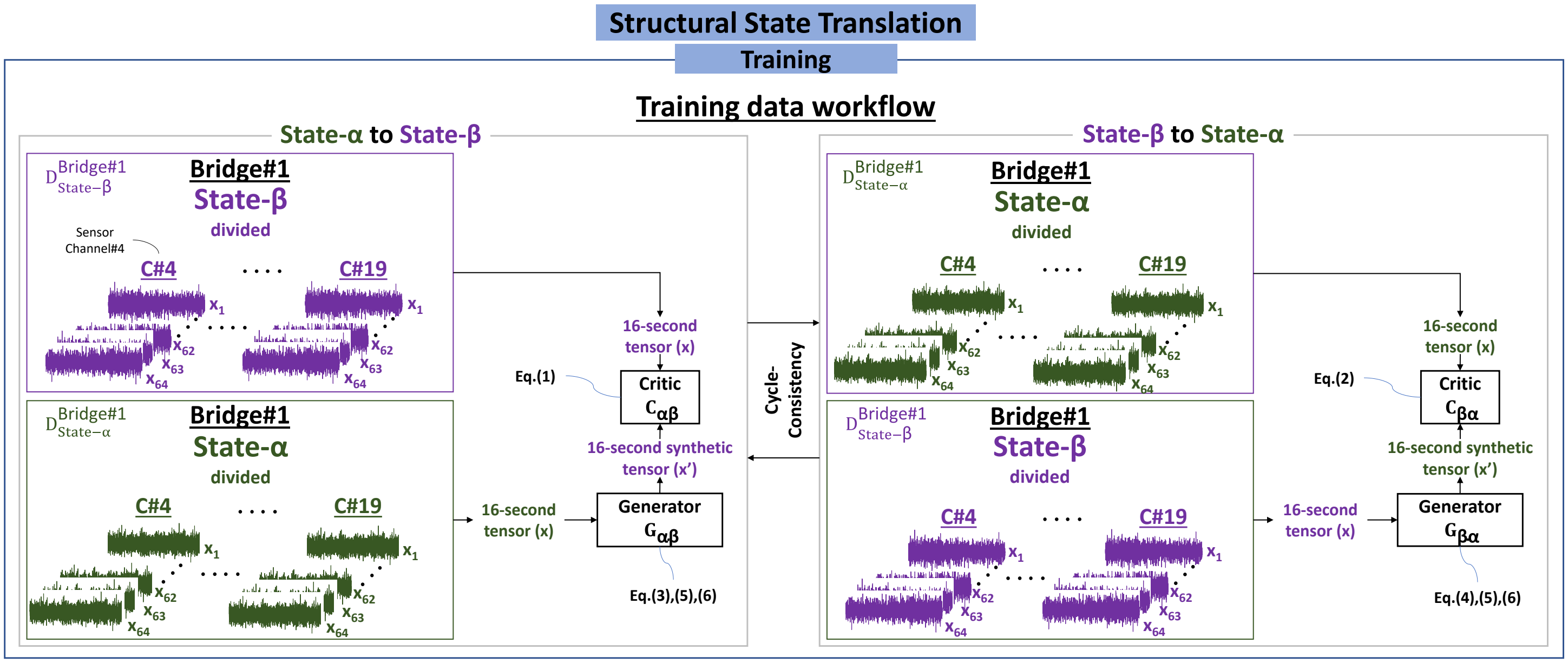

**Figure 11.** The training data workflow: the DGCG model is enforced to learn the domain invariancy between *State-α* and *State-β* of *Bridge#1* by training in a cycle-consistent manner on multi-domains of states, *State-α* and *State-β*.

$$Adversarial\ Critic\ Loss = \mathcal{L}_{DGCG}^{C_{\alpha\beta}}(G_{\alpha\beta}, C_{\alpha\beta}) = \mathbb{E}_{x\sim\mathbb{P}(X)}[C_{\alpha\beta}(x)]$$

$$-\mathbb{E}_{x\sim\mathbb{P}(X)}\left[C_{\alpha\beta}\left(G_{\alpha\beta}(x)\right)\right] + (\lambda_{GP})\mathbb{E}_{x'\sim\mathbb{P}(X')}\left[\left(\left\|\nabla_{x'}C_{\alpha\beta}(x')\right\|_2 - 1\right)^2\right] \quad (1)$$

$$Adversarial\ Critic\ Loss = \mathcal{L}_{DGCG}^{C_{\beta\alpha}}(G_{\beta\alpha}, C_{\beta\alpha}) = \mathbb{E}_{x\sim\mathbb{P}(X)}[C_{\beta\alpha}(x)]$$

$$-\mathbb{E}_{x\sim\mathbb{P}(X)}\left[C_{\beta\alpha}\left(G_{\beta\alpha}(x)\right)\right] + (\lambda_{GP})\mathbb{E}_{x'\sim\mathbb{P}(X')}\left[\left(\left\|\nabla_{x'}C_{\beta\alpha}(x')\right\|_2 - 1\right)^2\right] \quad (2)$$

$$Adversarial\ Generator\ Loss = \mathcal{L}_{DGCG}^{G_{\alpha\beta}}(G_{\alpha\beta}, C_{\alpha\beta}) = -\mathbb{E}_{x\sim\mathbb{P}(X)}\left[C_{\alpha\beta}\left(G_{\alpha\beta}(x)\right)\right] \quad (3)$$

$$Adversarial\ Generator\ Loss = \mathcal{L}_{DGCG}^{G_{\beta\alpha}}(G_{\beta\alpha}, C_{\beta\alpha}) = -\mathbb{E}_{x\sim\mathbb{P}(X)}\left[C_{\beta\alpha}\left(G_{\beta\alpha}(x)\right)\right] \quad (4)$$

$$Cycle\ Consistency\ Loss = \mathcal{L}_{DGCG}^{Cyc}(G_{\alpha\beta}, G_{\beta\alpha}) = \mathbb{E}_{x\sim\mathbb{P}(X)}\left[\left|G_{\beta\alpha}\left(G_{\alpha\beta}(x)\right) - x\right|_1\right] +$$

$$\mathbb{E}_{x\sim\mathbb{P}(X)}\left[\left|G_{\alpha\beta}\left(G_{\beta\alpha}(x)\right) - x\right|_1\right] \quad (5)$$

$$Identity\ Loss = \mathcal{L}_{DGCG}^{Id}(G_{\alpha\beta}, G_{\beta\alpha}) = \mathbb{E}_{x\sim\mathbb{P}(X)}\left[\left|G_{\alpha\beta}(x) - x\right|_1\right] + \mathbb{E}_{x\sim\mathbb{P}(X)}\left[\left|G_{\beta\alpha}(x) - x\right|_1\right] \quad (6)$$

$$\boldsymbol{Total\ Critic\ Losses} = \mathcal{L}_{DGCG}^{C_{\alpha\beta}}(G_{\alpha\beta}, C_{\alpha\beta}) + \mathcal{L}_{DGCG}^{C_{\beta\alpha}}(G_{\beta\alpha}, C_{\beta\alpha}) \quad (7)$$

$$\boldsymbol{Total\ Generator\ Losses} = \mathcal{L}_{DGCG}^{G_{\alpha\beta}}(G_{\alpha\beta}, C_{\alpha\beta}) + \mathcal{L}_{DGCG}^{G_{\beta\alpha}}(G_{\beta\alpha}, C_{\beta\alpha}) + \mathcal{L}_{DGCG}^{Cyc}(G_{\alpha\beta}, G_{\beta\alpha})\lambda_{Cyc} +$$

$$\mathcal{L}_{DGCG}^{Id}(G_{\alpha\beta}, G_{\beta\alpha})\lambda_{Id} \quad (8)$$

### 3.2.3. Evaluation Indices

It is pertinent to mention the indicators used to evaluate the accuracy of SST in this study. Fréchet Inception Distance (FID) [74] is one of the most used indices for evaluating GANs for image-based applications. Nevertheless, it is found that the FID is not quite sufficient for evaluating GAN for civil SHM applications since FID essentially considers the mean and variance values of the input data as shown in Eq. (9), where $\mu$ is the mean, $C$ is the covariance, and subscript $x$ denotes the original and subscript $x'$ denotes the generated data (or translated - synthetic). The acceleration response signals collected from civil structures, on the other hand, mean values of them are zero, and the variance remains mainly similar. This inefficiency of FID for evaluating GAN for 1-D SHM applications is also confirmed based on the observations made in a few studies, and more details can be seen in those studies [31–33,61]. A significant disadvantage of FID for similarity comparison is that it is intuitively challenging to understand the high or low FID value, as there are no upper or lower boundaries. On the other hand, analyzing the similarity between original and generated data in the frequency domain is more valuable as data analysis practices for SHM damage diagnosis and prognosis applications are critical and prevalent in the frequency domain. Therefore, a new indicator, Mean Magnitude-Squared Coherence (MMSC), is presented and used successfully [33,61].

$$FID(x,x') = ||\mu_x - \mu_{x'}||_2^2 + Tr\left(C_x + C_{x'} - 2(C_x C_{x'})^{1/2}\right) \tag{9}$$

MMSC is used to track the similarities of the original data $x$ and generated data $x'$ in their corresponding frequency domains. In Eq. (10), $S_{xx'}$ is the cross-spectral density estimate, and $S_{xx}$ and $S_{x'x'}$ are the power spectral density estimates of the original and the synthetic data, respectively. Following the computation of Magnitude-Squared Coherence (MSC) values of original and generated data in Eq. (10), the resulting $n$ amount of MSC values are averaged to give a single representative mean score as shown in Eq. (11). This mean score denotes the similarity of two data to each other in the frequency domain. Consequently, when the generated data is exactly the same as the original data in the frequency domain, the MMSC value is 1; when they are exactly dissimilar, it is 0. The FID values, on the other hand, the lower the value, the more similar the data samples are.

$$MSC(x,x') = \left[\frac{|S_{xx'}|^2}{S_{xx} \times S_{x'x'}}\right] \tag{10}$$

$$MMSC(x,x') = \frac{\sum_{i=1}^{n} MSC(x,x')_i}{n} \tag{11}$$

### 3.2.4. Evaluation of the Learning

The total critic and generator losses, FID, and MMSC values are monitored to track the learning of the DGCG model. In other words, the critic and generator losses and FID and MMSC values of the 16-second tensors in *State-α divided* ($D_{State-\alpha}^{Bridge\#1}$) between *State-$\hat{\alpha}$ divided* ($D_{State-\hat{\alpha}}^{Bridge\#1}$), and *State-β divided* ($D_{State-\beta}^{Bridge\#1}$) between *State-$\hat{\beta}$ divided* ($D_{State-\hat{\beta}}^{Bridge\#1}$) are calculated. The training procedure is summarized with an illustration in **Figure 12**.

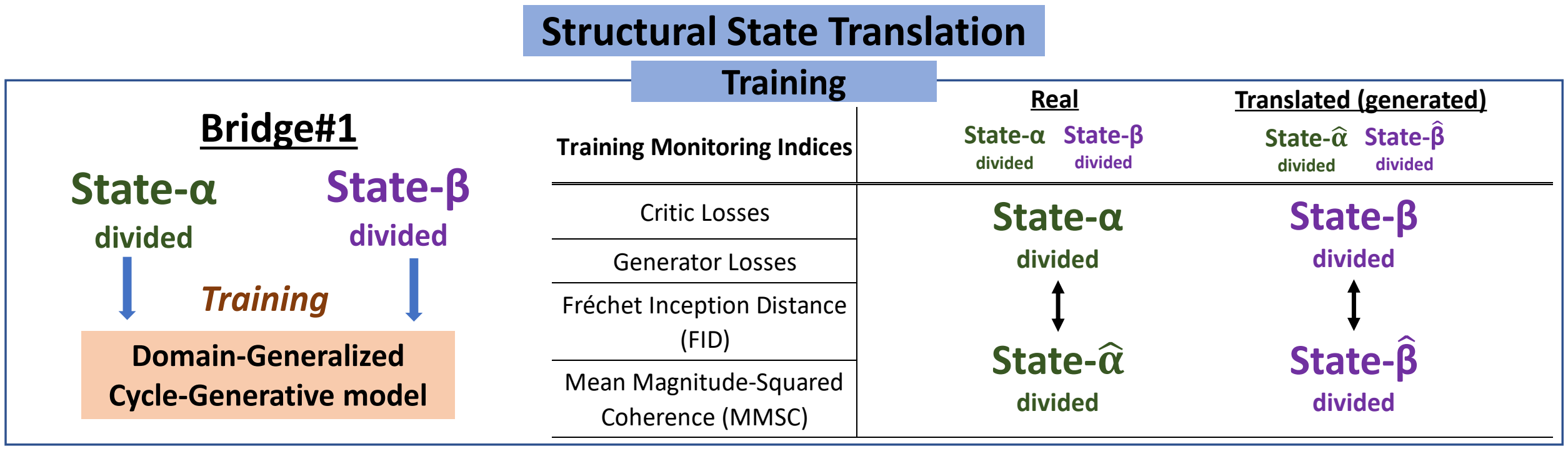

**Figure 12.** The summary of the training procedure

Next, the training monitoring indices are plotted in **Figure 13** over each iteration made in an epoch. It can be observed that the total generator losses, which is the monitored loss function of Eq. (8), converged to zero (**Figure 13 (a)**). On the other hand, the total critic losses, Eq. (7), did not completely converge to zero (**Figure 13 (b)**). With fine-tuning, better learning/training results could be achieved. The FID, Eq. (9), also converged to zero for both domains (**Figure 13 (c)-(d)**). However, several cases were tested in the previous experiments where the FID values followed similar trends as in this study, but the MMSC values were low, and the modal identification results of the generated datasets were not similar to the original datasets. The MMSC (Eq. (11)) values seem to be converged to 1 for both domains (**Figure 13 (e)-(f)**). Yet, the MMSC values in **Figure 13 (f)** look noisier than in **Figure 13 (e)**. Later in the study, this phenomenon reveals that the translated (generated or synthetic) states of *State-$\hat{\beta}$* and *State-$\hat{\gamma}$* are more similar to *State-β* and *State-γ* than *State-$\hat{\alpha}$* being similar to *State-α*. This fact indicates that the SST executed in *State-α* to *State-$\hat{\beta}$* and *State-α* to *State-$\hat{\gamma}$* are performed slightly better than the SST executed in *State-β* to *State-$\hat{\alpha}$* and *State-γ* to *State-$\hat{\alpha}$*. Overall, the training results show that the DGCG model has achieved a satisfactory learning process, particularly MMSC results indicating there is an almost one-to-one similarity between the generated and original data.

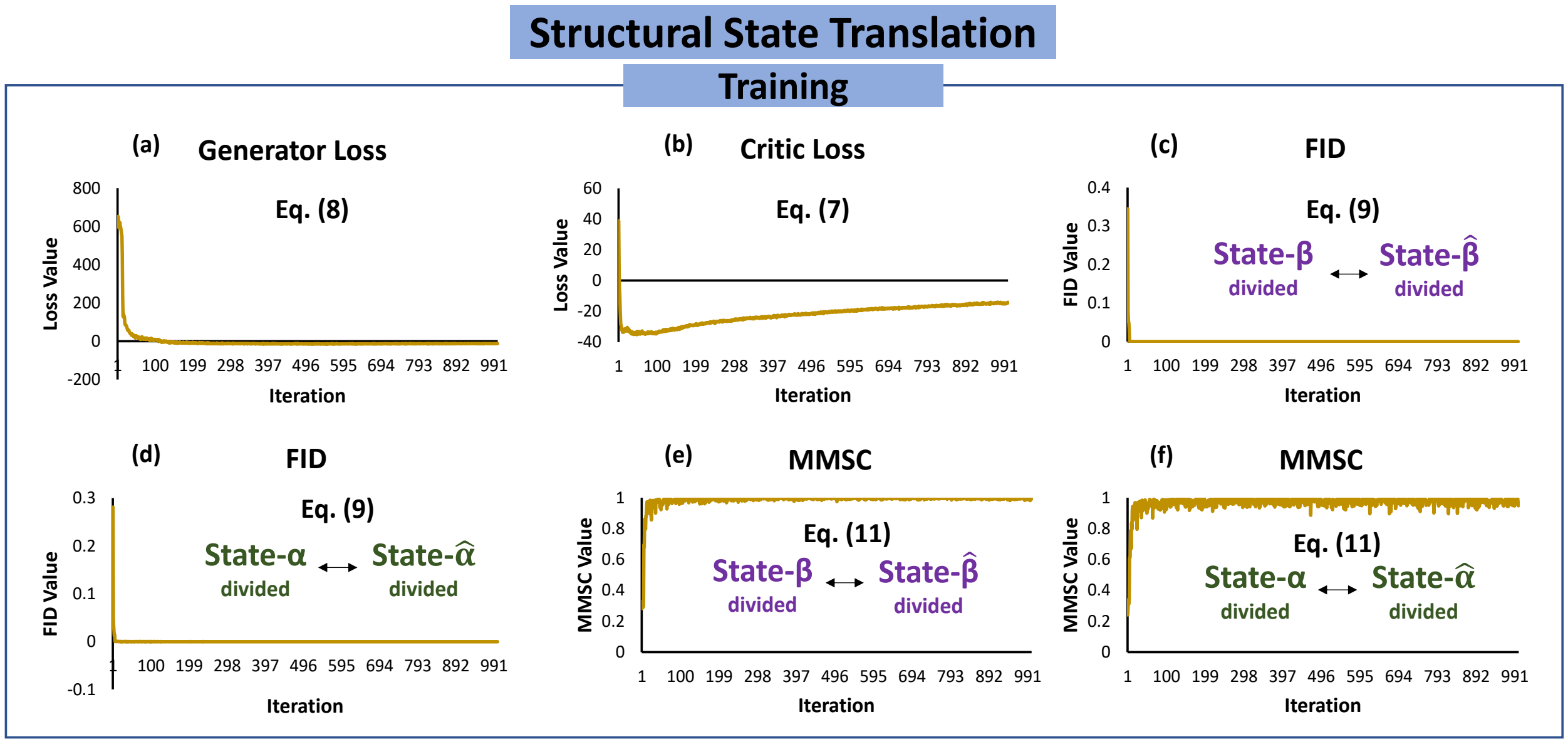

**Figure 13.** *Bridge#1* training monitoring plots over iterations: **(a)** Generator loss, **(b)** Critic loss, **(c)** FID values of *State-β divided* and *State-$\hat{\beta}$ divided*, **(d)** FID values of *State-α divided* and *State-$\hat{\alpha}$ divided*, **(e)** MMSC values of *State-β divided* and *State-$\hat{\beta}$ divided*, **(f)** MMSC values of *State-α divided* and *State-$\hat{\alpha}$ divided*.

### 3.3. Translation

In essence, what is translated in the Translation phase is the domain of each bridge state, a data domain translation. The Translation phase simply consists of having the DGCG model translate the divided states to other states, as shown in Phase 3 of **Figure 8**. After the DGCG model has trained on the domains $D_{State-\alpha}^{Bridge\#1}$ (*State-α divided*) and $D_{State-\beta}^{Bridge\#1}$ (*State-β divided*), the model is used to translate *State-α divided* to *State-$\hat{\beta}$ divided*, *State-α divided* to *State-$\hat{\gamma}$ divided*, *State-β divided* to *State-$\hat{\alpha}$ divided*, and lastly, *State-γ divided* to *State-$\hat{\alpha}$ divided* for each *Bridge#2*, *Bridge#3*, and *Bridge#4*, as illustrated in **Figure 3**. Note that "^" denotes the state is the synthetic (generated/translated) data, as mentioned previously. Additionally, as aforementioned, *State-β* and *State-γ* are structurally symmetrical. Thus, the translated *State-$\hat{\beta}$* from *State-α* and the translated *State-$\hat{\gamma}$* from *State-α* should be the same in terms of their structural (modal) parameters. Conversely, the translated *State-$\hat{\alpha}$* from *State- β* and the translated *State-$\hat{\alpha}$* from *State-γ* should be the same.

Moreover, each SST procedure for each bridge is carried out under separate scenarios. For instance, in Scenario I, *Bridge#2* is assumed to only have data for the *State-α* condition, and the aim is to make the data available for another condition, *State-β*, which is the removal of the bottom chord of the bridge. Similarly, in Scenario II, *Bridge#2* is assumed to only have data for the *State-β* condition, and the aim is to make the data available for another condition, *State-a*, which is the pristine condition of the bridge. The other scenarios are produced in a similar fashion, as shown in **Figure 3** and **Figure 8**.

### 3.4. Postprocessing

After the Translation phase, the Postprocessing phase is carried out, where the reverse process of Preprocessing phase is implemented. Essentially, the translated 16-second tensors in the domains of each *State-$\hat{\alpha}$ divided, State-$\hat{\beta}$ divided*, and *State-$\hat{\gamma}$ divided*, are concatenated to generate the 1024-second signals per sensor channel for each state of each bridge. The preprocessing phase is illustrated in **Figure 14**.

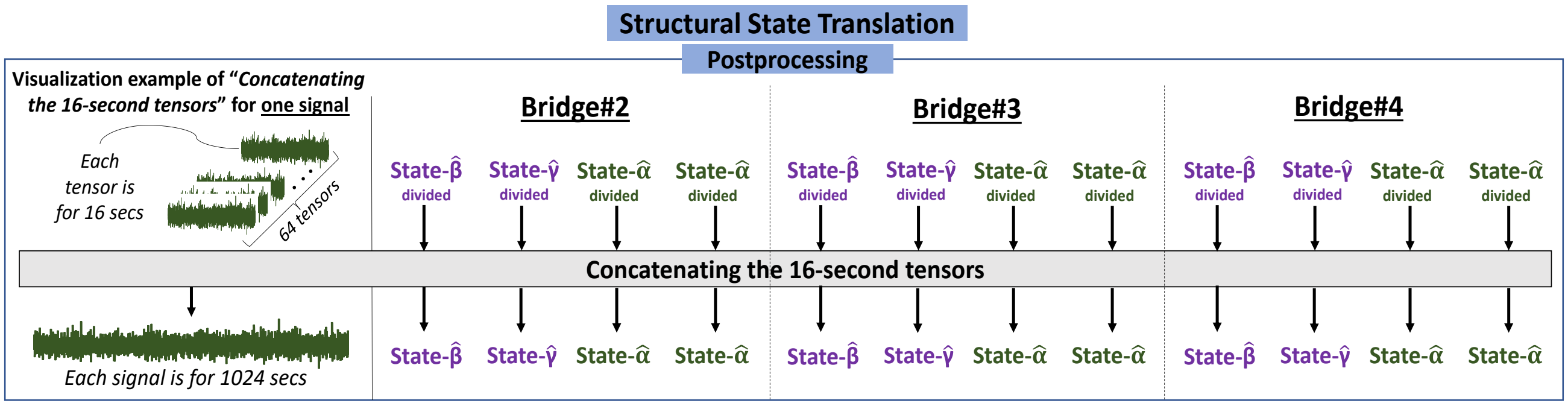

**Figure 14.** Postprocessing: the reverse process of the Preprocessing phase is carried out where the concatenation of the 16-second tensors in the domains of each state is taken place.

## 4. Evaluation of the Results of Structural State Translation

After concatenating the 16-second synthetic tensors to form the full signals in the states of each bridge, the evaluation of each target bridge's translated states is investigated. First, the evaluation is done using the MMSC index for each state of each target bridge. For that, the MMSC values are computed between the signal pairs from each sensor channel of real and synthetic states of each target bridge. Note that there are two *State-$\hat{\alpha}$*, which were translated from *State-$\beta$* and *State-$\gamma$*. Thus, to avoid confusion, the states are represented with alphabetic letters, as shown in **Figure 15** for *Bridge#2*. In **Figure 15**, the MMSC values are computed between the signal pairs from each channel of *State-$\alpha$* and *State-$\hat{\alpha}$* ((a) – (b)), *State-$\alpha$* and *State-$\hat{\alpha}$* ((c) – (d)), *State-$\beta$* and *State-$\hat{\beta}$* ((e) – (f)), and *State- $\gamma$* and *State-$\hat{\gamma}$* ((g) – (h)). In addition, the MMSC values are plotted, showing the values throughout the sensor channels on *Bridge#2*. The same evaluation approach is implemented for *Bridge#3* and *Bridge#4*, which are shown in **Figure 16-17**, respectively. As can be seen from the figures, while the MMSC values are generally above 0.90s, some values go down to 0.84s for *Bridge#2*, meaning that the minimum similarity between real and synthetic signals from each sensor channel starts from 84%. Whereas for the other bridges, the MMSC values are much higher. Additionally, the plots of MMSC values through the channels reveal that the MMSC values are roughly the same between one-half of the bridge and the other half, which is understandable because the bridges are structurally symmetrical. A more detailed discussion of the evaluation is made after the modal identification process of the real and synthetic (translated) states in the following paragraphs.

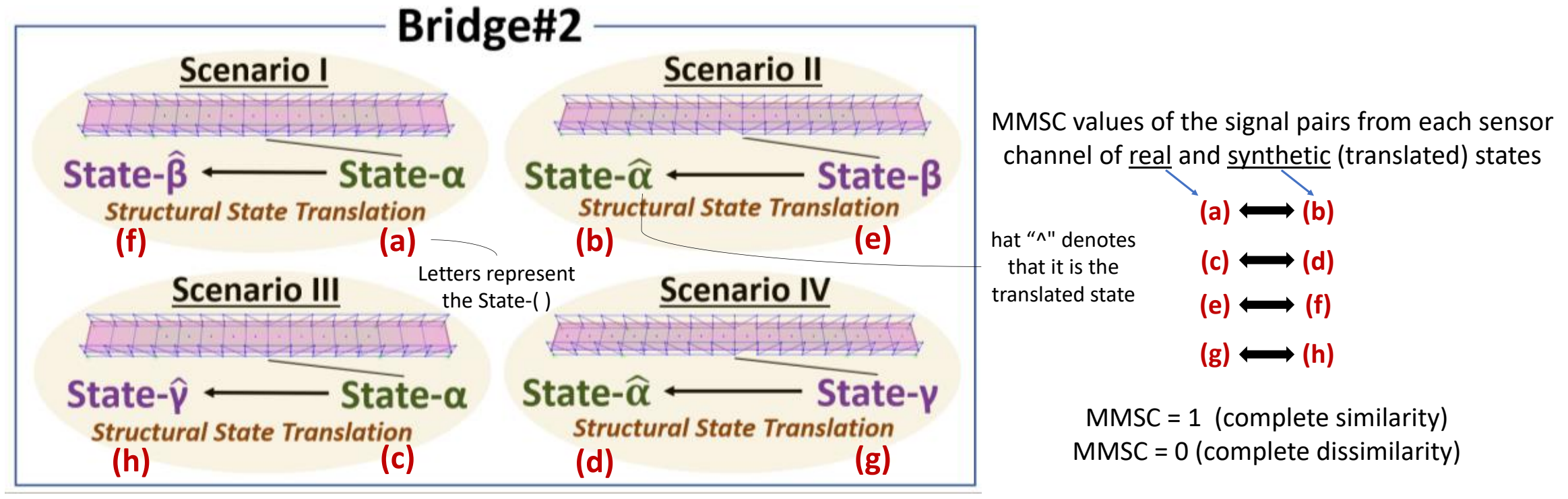

**MMSC values**

| | (a) ⟷ (b) | (c) ⟷ (d) | (e) ⟷ (f) | (g) ⟷ (h) |
|---|---|---|---|---|
| Channel#7 | 0.934 | 0.937 | 0.941 | 0.941 |
| Channel#8 | 0.948 | 0.933 | 0.939 | 0.934 |
| Channel#9 | 0.934 | 0.908 | 0.947 | 0.929 |
| Channel#10 | 0.884 | 0.918 | 0.932 | 0.935 |
| Channel#11 | 0.843 | 0.889 | 0.912 | 0.923 |
| Channel#12 | 0.887 | 0.843 | 0.922 | 0.912 |
| Channel#13 | 0.916 | 0.882 | 0.934 | 0.932 |
| Channel#14 | 0.908 | 0.934 | 0.929 | 0.948 |
| Channel#15 | 0.935 | 0.948 | 0.936 | 0.942 |
| Channel#16 | 0.936 | 0.931 | 0.939 | 0.939 |
| Average MMSC of the whole bridge | 0.912 | 0.912 | 0.933 | 0.933 |

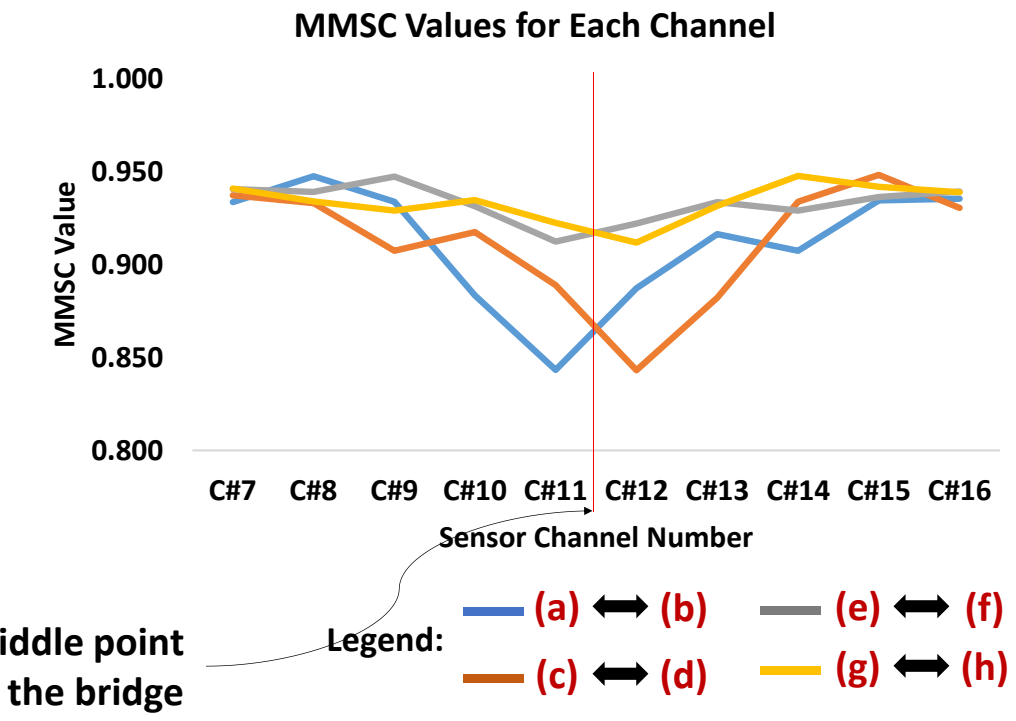

**Figure 15.** MMSC values of the signal pairs from each sensor channel of real and translated states of *Bridge#2*.

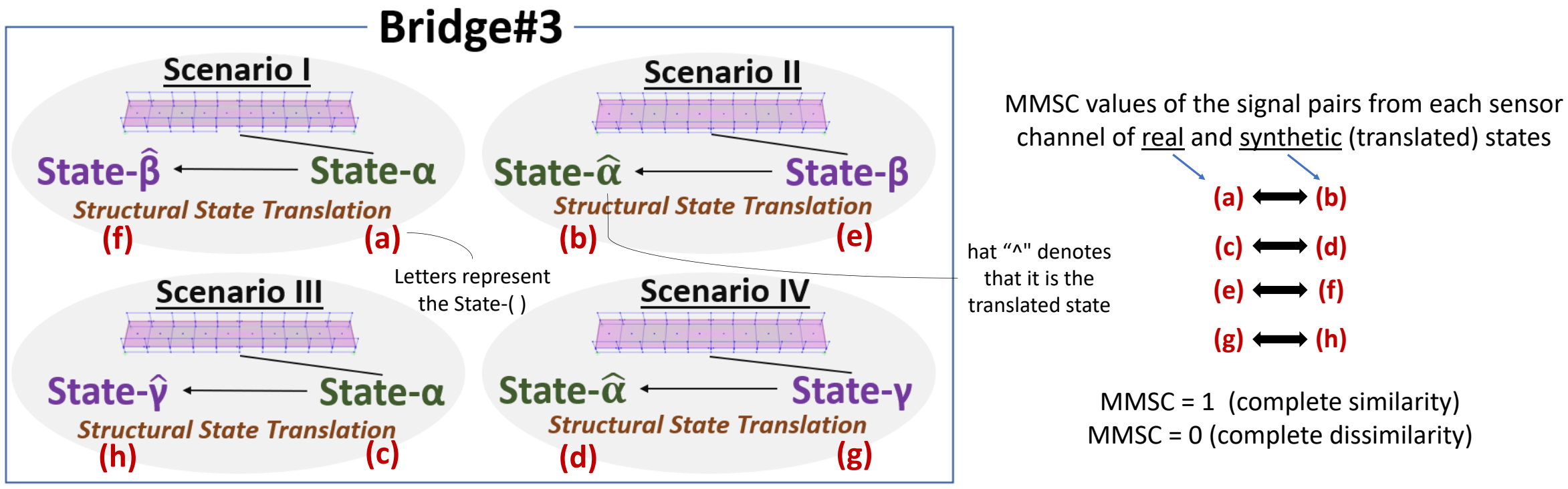

**MMSC values**

| | (a) ⟷ (b) | (c) ⟷ (d) | (e) ⟷ (f) | (g) ⟷ (h) |
|---|---|---|---|---|
| Channel#7 | 0.938 | 0.937 | 0.943 | 0.942 |
| Channel#8 | 0.957 | 0.957 | 0.979 | 0.979 |
| Channel#9 | 0.962 | 0.963 | 0.980 | 0.981 |
| Channel#10 | 0.966 | 0.967 | 0.979 | 0.979 |
| Channel#11 | 0.960 | 0.961 | 0.970 | 0.972 |
| Channel#12 | 0.959 | 0.959 | 0.970 | 0.969 |
| Channel#13 | 0.967 | 0.967 | 0.979 | 0.979 |
| Channel#14 | 0.963 | 0.962 | 0.981 | 0.980 |
| Channel#15 | 0.957 | 0.957 | 0.979 | 0.979 |
| Channel#16 | 0.940 | 0.941 | 0.948 | 0.949 |
| Average MMSC of the whole bridge | 0.957 | 0.957 | 0.971 | 0.971 |

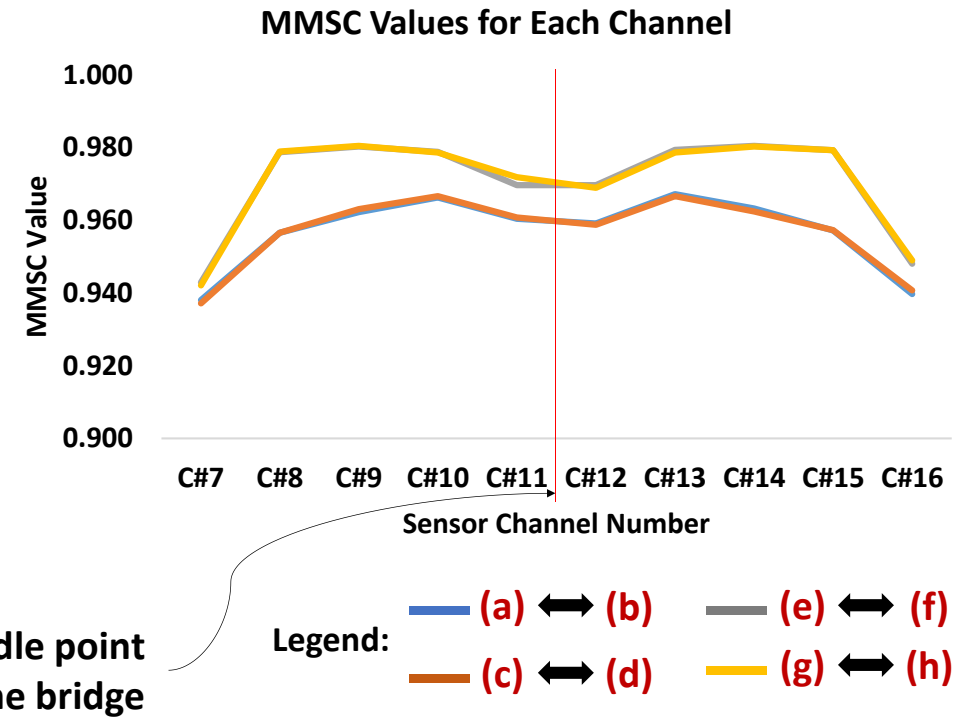

**Figure 16.** MMSC values of the signal pairs from each sensor channel of real and translated states of *Bridge#3*.

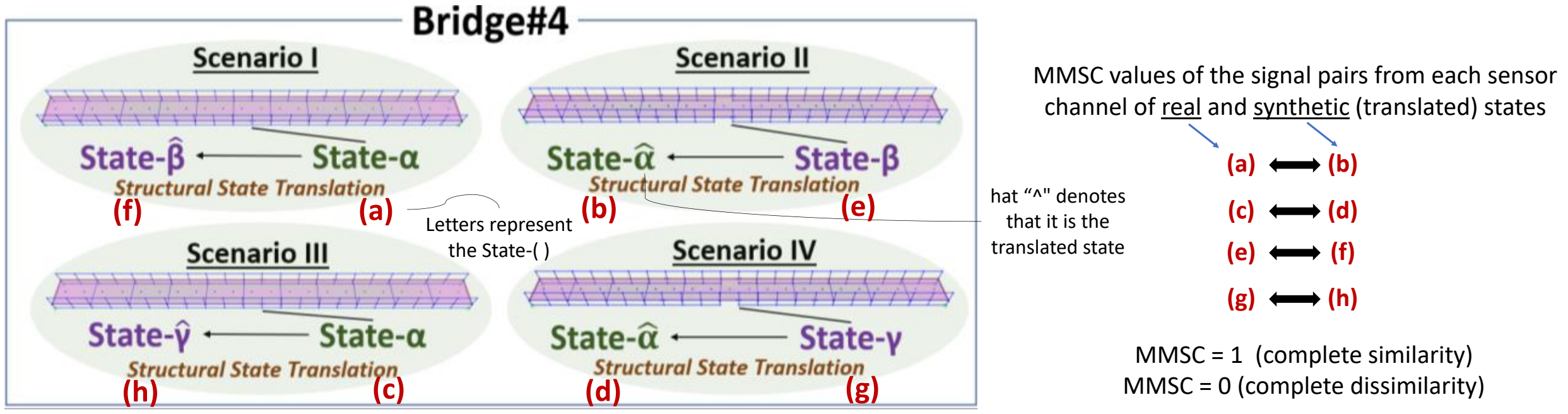

**MMSC values**

| | (a) ⟷ (b) | (c) ⟷ (d) | (e) ⟷ (f) | (g) ⟷ (h) |
|---|---|---|---|---|
| Channel#1 | 0.938 | 0.936 | 0.942 | 0.942 |
| Channel#2 | 0.960 | 0.961 | 0.961 | 0.961 |
| Channel#3 | 0.966 | 0.967 | 0.958 | 0.958 |
| Channel#4 | 0.959 | 0.970 | 0.972 | 0.959 |
| Channel#5 | 0.972 | 0.971 | 0.955 | 0.955 |
| Channel#6 | 0.968 | 0.968 | 0.947 | 0.946 |
| Channel#7 | 0.965 | 0.964 | 0.948 | 0.947 |
| Channel#8 | 0.958 | 0.958 | 0.963 | 0.963 |
| Channel#9 | 0.953 | 0.954 | 0.966 | 0.966 |
| Channel#10 | 0.943 | 0.943 | 0.967 | 0.966 |
| Channel#11 | 0.933 | 0.933 | 0.964 | 0.964 |
| Channel#12 | 0.931 | 0.932 | 0.963 | 0.963 |
| Channel#13 | 0.942 | 0.942 | 0.966 | 0.966 |
| Channel#14 | 0.955 | 0.954 | 0.967 | 0.967 |
| Channel#15 | 0.960 | 0.960 | 0.964 | 0.964 |
| Channel#16 | 0.965 | 0.965 | 0.946 | 0.947 |
| Channel#17 | 0.969 | 0.969 | 0.947 | 0.947 |
| Channel#18 | 0.972 | 0.972 | 0.955 | 0.955 |
| Channel#19 | 0.971 | 0.971 | 0.957 | 0.956 |
| Channel#20 | 0.968 | 0.966 | 0.960 | 0.960 |
| Channel#21 | 0.961 | 0.960 | 0.962 | 0.962 |
| Channel#22 | 0.937 | 0.941 | 0.941 | 0.938 |
| Average MMSC of the whole bridge | 0.957 | 0.957 | 0.958 | 0.957 |

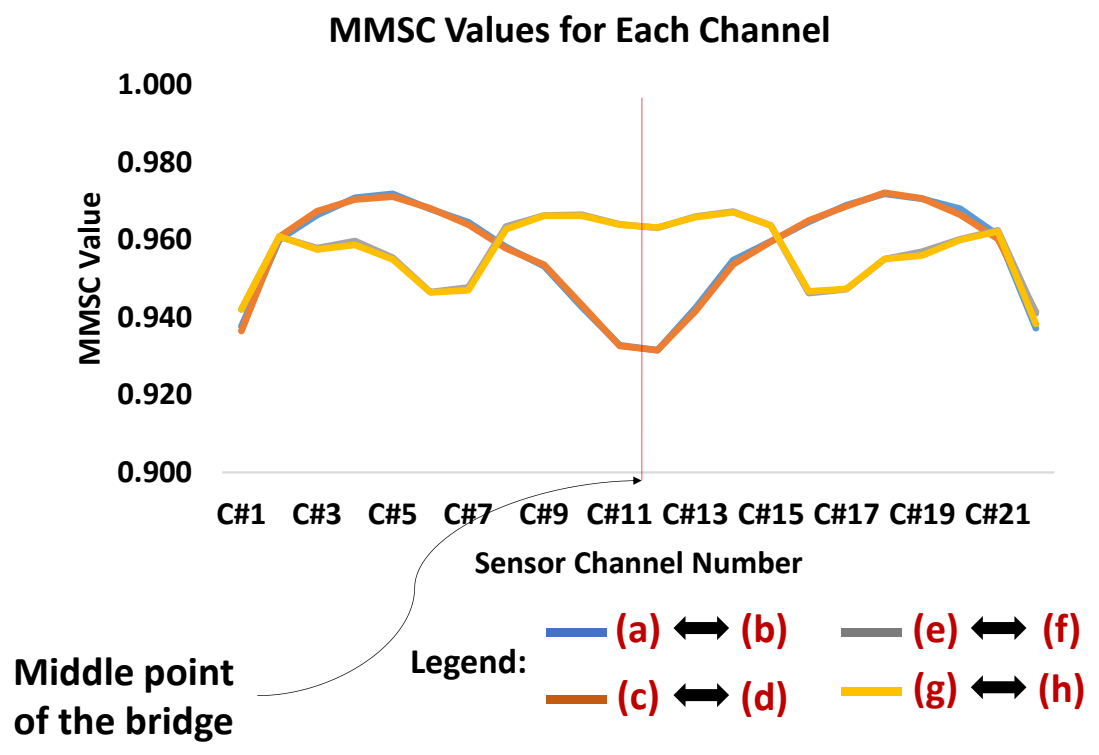

**Figure 17.** MMSC values of the signal pairs from each sensor channel of real and translated states of *Bridge#4*.

While comparing the real and translated states via MMSC may give intuition about the signals' similarities, understanding each bridge state's physical meaning is critical in the SHM context. Therefore, a modal identification process is implemented for the state of each bridge. First, the geometry of the bridges used for the modal identification is modelled, as shown in **Figure 18**. Then, the Frequency Domain Decomposition (FDD) method [75] is used with 66% Hann window overlapping and a resolution of 1024 frequency lines for each bridge state. Subsequently, the modes are identified by the peak picking technique from the obtained singular values of power spectral densities and accordingly, mode shapes and natural frequencies are extracted. In **Figure 19-21**, the modal information of the bridge states is given in a similar fashion to **Figure 15-17**. Additionally, for each state comparison of each bridge, the Change in Natural Frequencies % (CNF) and Modal Assurance Criterion (MAC) for the dominating modes that were picked are calculated and compared, as well as the illustrations of the mode shapes provided. Lastly, the average

MMSC values of each state comparison are shown as they are related to the modal identification results, which is discussed in the following paragraphs.

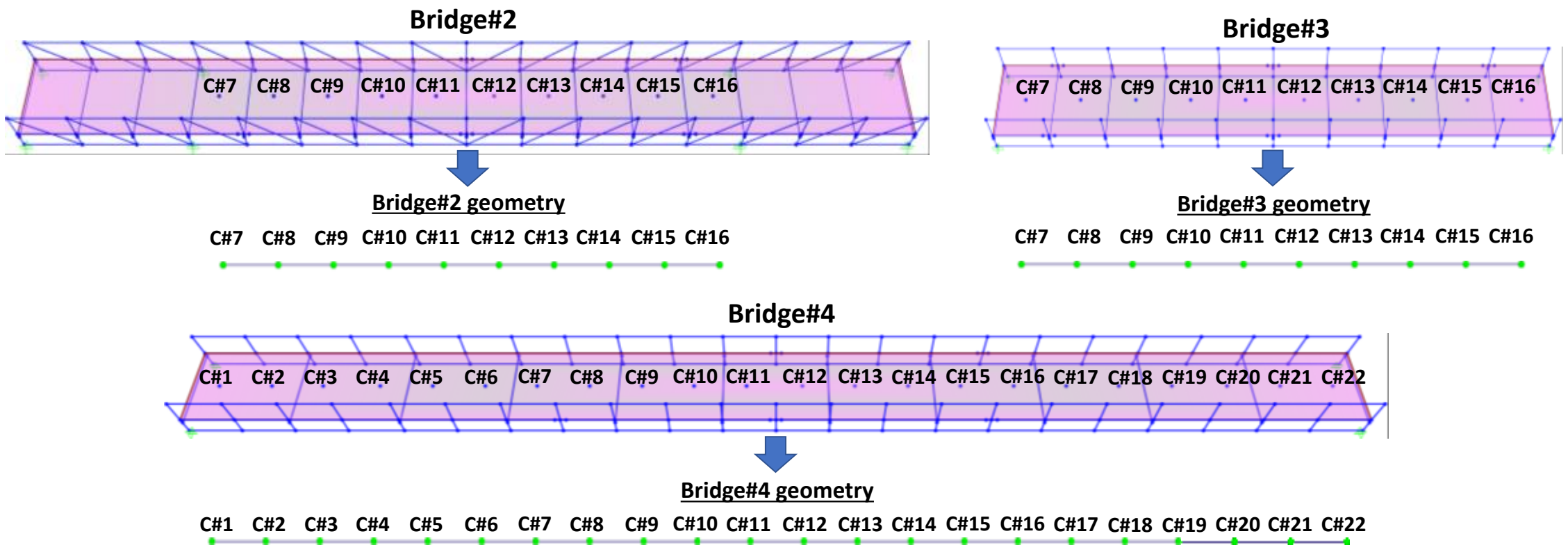

**Figure 18.** The bridge geometries used for the modal identification procedure on the extracted signals from the bridges.

Overall, it can be observed from **Figure 14-16** and **Figure 19-21** that the DGCG model achieved successful SST performance in each scenario. The MMSC values demonstrated that the translated bridge states are extremely similar to the real states. As such, from the comparison of each bridge state, the lowest and the highest average MMSC values are observed to be 91.2% and 97.1%, respectively. Moreover, the natural frequencies and the mode shapes between each bridge state are also observed to be significantly similar. Hence, the highest and the lowest difference in natural frequencies among the modes of the bridge states are, respectively, 5.71% and 0%, while the highest and lowest MAC values are 0.998 and 0.87. Accordingly, some observations and interpretations are made, which are presented in the following paragraphs.

**Side note.** The modes of each bridge state using Modal Analysis in the FEA program account for torsional, lateral, and longitudinal modes. However, the modal identification process implemented via FDD on the extracted signals from bridges accounts for bending modes due to the layout of virtual sensors on the models (single-line layout). Therefore, the torsional, lateral, and longitudinal modes were not visible in the identified modes. Some other bending modes were also not detected in the singular values of power spectral densities, particularly for *Bridge#2* due to being the stiffest bridge among other bridges. In addition, considering the typical numerical errors and possible noise interruption during the data extraction in the FEA programs, the number of dominating modes identified on the extracted data from the bridges was lower than the modes obtained numerically in FEA. As a result, 2 modes for *Bridge#2*, 4 modes for *Bridge#3*, and 6 modes for *Bridge#4* could be identified. This makes sense as the overall stiffness of the bridges could be ranked in ascending order as *Bridge#2*, *Bridge#3*, and *Bridge#4* (**Figure 7**), which is a similar order to the number of dominating modes that could be identified. Nonetheless, the identified modes after FDD show that the generated modes are significantly similar to the real modes (**Figure 19-21**).

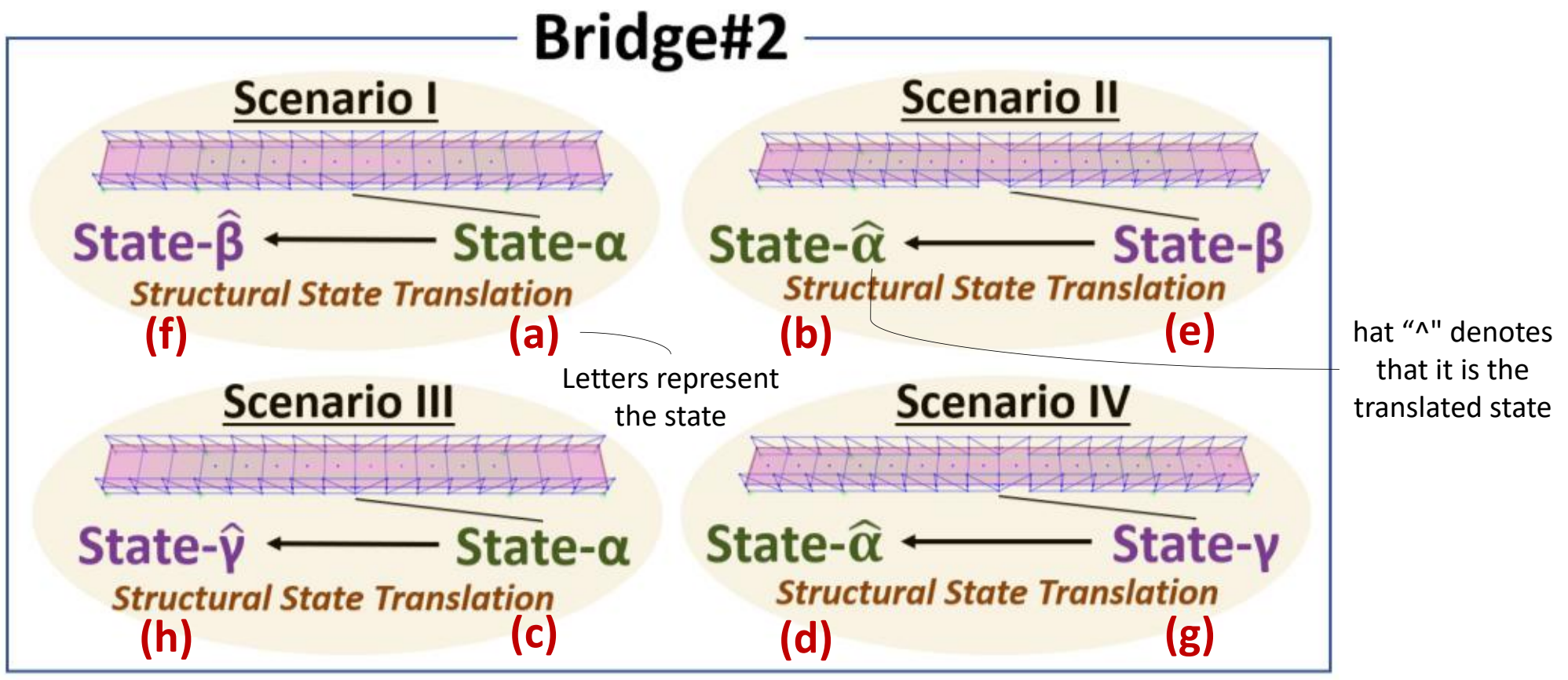

(a) ⟷ (b) Avg. MMSC = 0.912

| Mode | State-α Natural Frequencies (Hz) | State-α̂ Natural Frequencies (Hz) | Change in Natural Frequencies (%) | Modal Assurance Criterion |
|---|---|---|---|---|
| 1 | 7.208 | 7.083 | 1.76 | 0.969 |
| 2 | 12.375 | 12.375 | 0.00 | 0.972 |

(c) ⟷ (d) Avg. MMSC = 0.912

| Mode | State-α Natural Frequencies (Hz) | State-α̂ Natural Frequencies (Hz) | Change in Natural Frequencies (%) | Modal Assurance Criterion |
|---|---|---|---|---|
| 1 | 7.208 | 7.083 | 1.76 | 0.972 |
| 2 | 12.375 | 12.375 | 0.00 | 0.963 |

(e) ⟷ (f) Avg. MMSC = 0.933

| Mode | State-β Natural Frequencies (Hz) | State-β̂ Natural Frequencies (Hz) | Change in Natural Frequencies (%) | Modal Assurance Criterion |
|---|---|---|---|---|
| 1 | 7.083 | 7.208 | 1.73 | 0.990 |
| 2 | 12.375 | 12.375 | 0.00 | 0.956 |

(g) ⟷ (h) Avg. MMSC = 0.933

| Mode | State-γ Natural Frequencies (Hz) | State-γ̂ Natural Frequencies (Hz) | Change in Natural Frequencies (%) | Modal Assurance Criterion |
|---|---|---|---|---|
| 1 | 7.083 | 7.208 | 1.73 | 0.989 |
| 2 | 12.375 | 12.375 | 0.00 | 0.956 |

**Figure 19.** The comparison of mode shapes and natural frequencies between each real and translated states of *Bridge#2*.

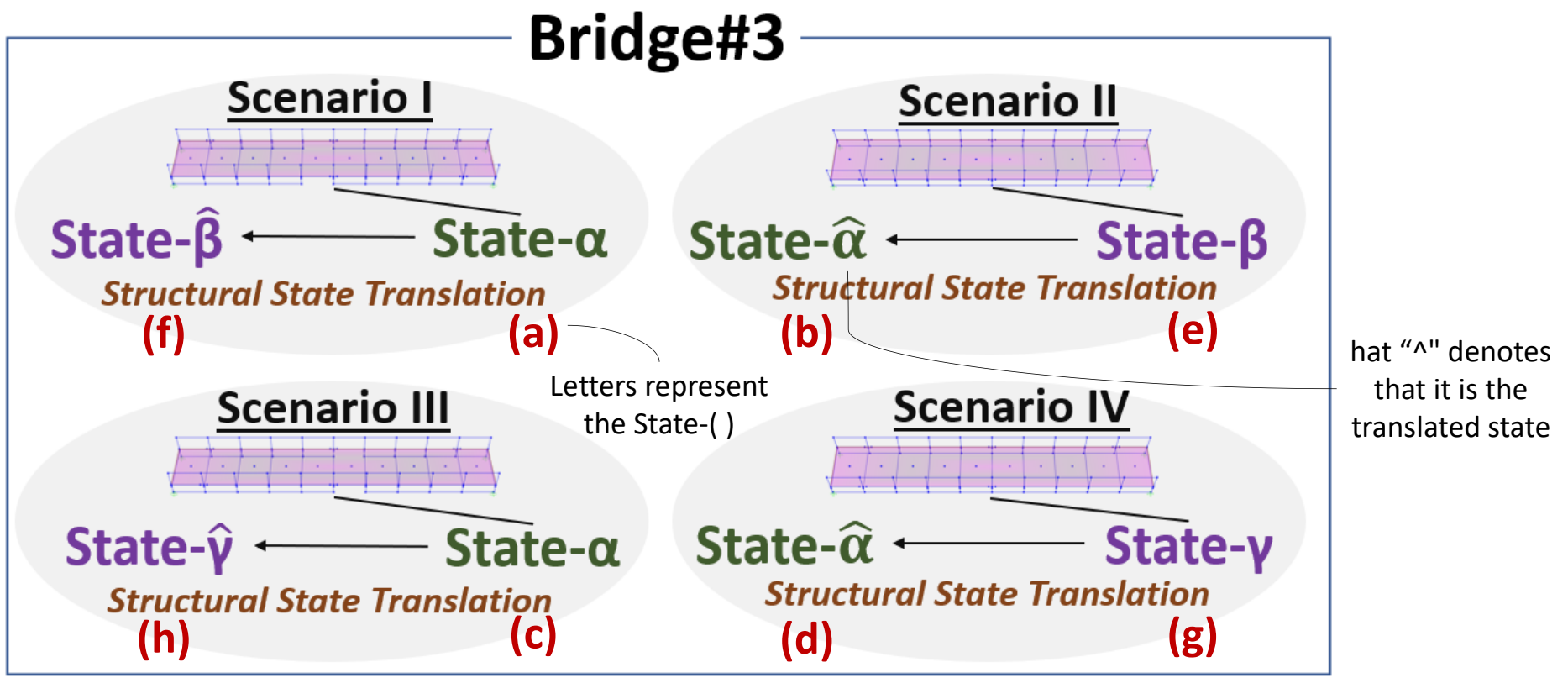

(a) ⟷ (b) Avg. MMSC = 0.957

| Mode | State-α Natural Frequencies (Hz) | State-α̂ Natural Frequencies (Hz) | Change in Natural Frequencies (%) | Modal Assurance Criterion |
|---|---|---|---|---|
| 1 | 2.042 | 2.042 | 0.00 | 0.995 |
| 2 | 7.083 | 7.083 | 0.00 | 0.870 |
| 3 | 14.167 | 13.917 | 1.80 | 0.967 |
| 4 | 22.167 | 22.167 | 0.00 | 0.981 |

(c) ⟷ (d) Avg. MMSC = 0.957

| Mode | State-α Natural Frequencies (Hz) | State-α̂ Natural Frequencies (Hz) | Change in Natural Frequencies (%) | Modal Assurance Criterion |
|---|---|---|---|---|
| 1 | 2.042 | 2.042 | 0.00 | 0.995 |
| 2 | 7.083 | 7.083 | 0.00 | 0.871 |
| 3 | 14.167 | 13.917 | 1.80 | 0.969 |
| 4 | 22.167 | 22.167 | 0.00 | 0.981 |

(e) ⟷ (f) Avg. MMSC = 0.971

| Mode | State-β Natural Frequencies (Hz) | State-β̂ Natural Frequencies (Hz) | Change in Natural Frequencies (%) | Modal Assurance Criterion |
|---|---|---|---|---|
| 1 | 2.042 | 2.042 | 0.00 | 0.998 |
| 2 | 7.083 | 7.083 | 0.00 | 0.983 |
| 3 | 14.167 | 14.167 | 0.00 | 0.997 |
| 4 | 22.167 | 22.167 | 0.00 | 0.998 |

(g) ⟷ (h) Avg. MMSC = 0.971

| Mode | State-γ Natural Frequencies (Hz) | State-γ̂ Natural Frequencies (Hz) | Change in Natural Frequencies (%) | Modal Assurance Criterion |
|---|---|---|---|---|
| 1 | 2.042 | 2.042 | 0.00 | 0.998 |
| 2 | 7.083 | 7.083 | 0.00 | 0.983 |
| 3 | 14.167 | 14.167 | 0.00 | 0.997 |
| 4 | 22.167 | 22.167 | 0.00 | 0.997 |

**Figure 20.** The comparison of mode shapes and natural frequencies between each real and translated states of *Bridge#3*.

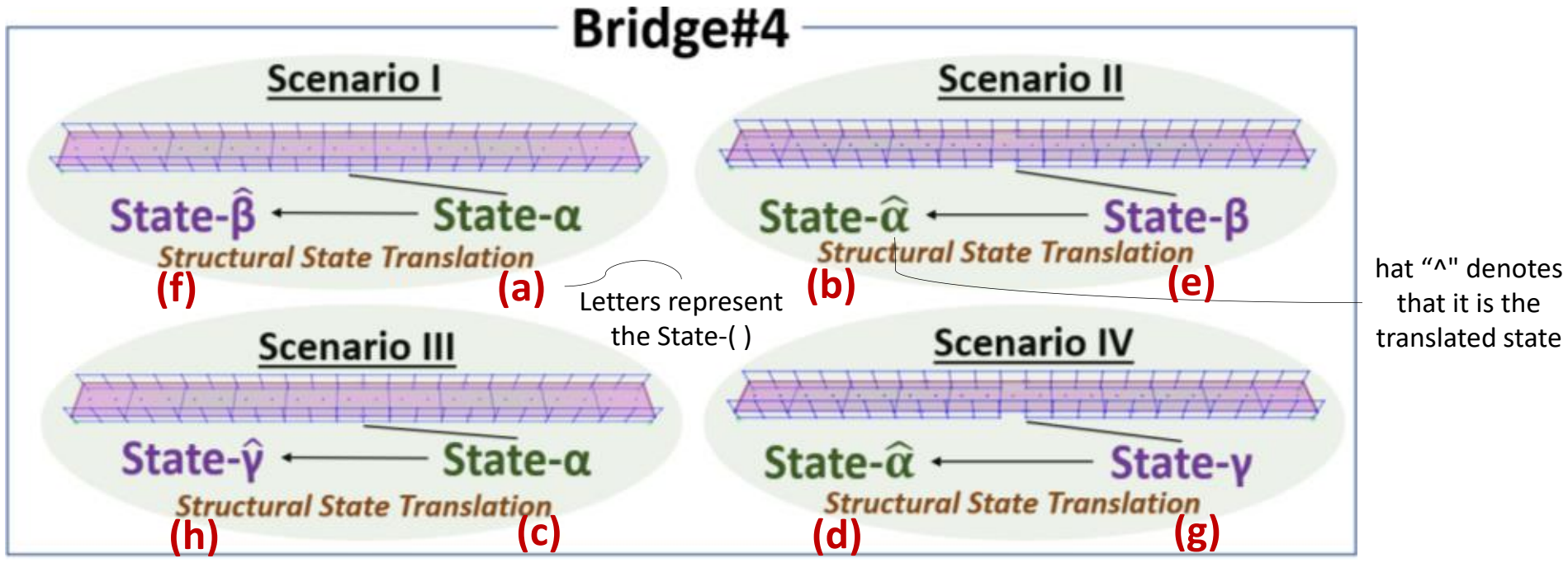

(a) ⟷ (b) Avg. MMSC = 0.957

| Mode | State-α (Natural Frequencies (Hz)) | State-α̂ (Natural Frequencies (Hz)) | Change in Natural Frequencies (%) | Modal Assurance Criterion |
|---|---|---|---|---|
| 1 | 1.083 | 1.083 | 0.00 | 0.982 |
| 2 | 3.833 | 3.833 | 0.00 | 0.960 |
| 3 | 6.583 | 6.583 | 0.00 | 0.969 |
| 4 | 10.792 | 10.792 | 0.00 | 0.953 |
| 5 | 15.375 | 14.542 | 5.73 | 0.913 |
| 6 | 26.157 | 25.417 | 2.91 | 0.961 |

(c) ⟷ (d) Avg. MMSC = 0.957

| Mode | State-α (Natural Frequencies (Hz)) | State-α̂ (Natural Frequencies (Hz)) | Change in Natural Frequencies (%) | Modal Assurance Criterion |
|---|---|---|---|---|
| 1 | 1.083 | 1.083 | 0.00 | 0.982 |
| 2 | 3.833 | 3.833 | 0.00 | 0.959 |
| 3 | 6.583 | 6.583 | 0.00 | 0.969 |
| 4 | 10.792 | 10.792 | 0.00 | 0.954 |
| 5 | 15.375 | 14.542 | 5.73 | 0.913 |
| 6 | 26.157 | 25.417 | 2.91 | 0.961 |

(e) ⟷ (f) Avg. MMSC = 0.958

| Mode | State-β (Natural Frequencies (Hz)) | State-β̂ (Natural Frequencies (Hz)) | Change in Natural Frequencies (%) | Modal Assurance Criterion |
|---|---|---|---|---|
| 1 | 1.083 | 1.083 | 0.00 | 0.910 |
| 2 | 3.542 | 3.542 | 0.00 | 0.996 |
| 3 | 6.583 | 6.583 | 0.00 | 0.970 |
| 4 | 10.792 | 10.792 | 0.00 | 0.966 |
| 5 | 14.542 | 15.375 | 5.42 | 0.979 |
| 6 | 25.417 | 26.167 | 2.87 | 0.983 |

(g) ⟷ (h) Avg. MMSC = 0.957

| Mode | State-γ (Natural Frequencies (Hz)) | State-γ̂ (Natural Frequencies (Hz)) | Change in Natural Frequencies (%) | Modal Assurance Criterion |
|---|---|---|---|---|
| 1 | 1.083 | 1.083 | 0.00 | 0.909 |
| 2 | 3.542 | 3.542 | 0.00 | 0.991 |
| 3 | 6.583 | 6.583 | 0.00 | 0.970 |
| 4 | 10.792 | 10.792 | 0.00 | 0.965 |
| 5 | 14.542 | 15.375 | 5.42 | 0.979 |
| 6 | 25.417 | 26.167 | 2.87 | 0.983 |

**Figure 21.** The comparison of mode shapes and natural frequencies between each real and translated states of *Bridge#4*.

**Observation 1.** Generally, the SSTs executed in Scenario I and Scenario III in **Figure 19-21** are slightly better. As such, the values in Percent Change in Natural Frequency (CNF) between *State-$\beta$* and *State-$\hat{\beta}$*, and *State-$\gamma$* and *State-$\hat{\gamma}$*, are slightly lower than the comparisons of both *State-$\alpha$* and *State-$\hat{\alpha}$*, which the SSTs for them were carried out in Scenario II and Scenario IV. Similarly, the Modal Assurance Criteria (MAC) and average MMSC values between *State-$\beta$* and *State-$\hat{\beta}$*, and *State-$\gamma$* and *State-$\hat{\gamma}$*, are slightly higher than the comparisons of both *State-$\alpha$* and *State-$\hat{\alpha}$*. This observation indicates that the SST process was slightly more successful in generating unhealthy (damaged) states, which was initially observed during the training of the DGCG model, where the MMSC values in **Figure 13 (f)** looked noisier than in **Figure 13 (e)**.

**Observation 2.** The SST evaluation results for *Bridge#3* (**Figure 16** and **Figure 20**) are slightly better than those of other bridges. As such, the CNF values are lower, and MAC and average MMSC values are higher for *Bridge#3*. This is understandable as the stiffness/flexibility of Bridge#1, which the model DGCG was trained with, is most similar to the stiffness/flexibility of *Bridge#3*. In other words, since *Bridge#1* and *Bridge#3* are most similar to each other in terms of their structural parameters, it makes sense when DGCG performs better in the scenarios of *Bridge#3* than the scenarios of other target bridges.

**Observation 3.** When there is a new mode appearance/mode switch in the states of *Bridge#2*, *Bridge#3*, and *Bridge#4* (**Table 2-3**), those particular modes in the translated states are slightly off than the ones in real states, which are confirmed with small values in CNF and slightly lower MAC values. This is because the new mode appearance/mode switch in the target/test domain (the states of *Bridge#2*, *Bridge#3*, and *Bridge#4*) is different from what the DGCG model already knows from the source/training domain (the states of *Bridge#1*). As mentioned, the new mode appearance/mode switch is observed when the states of the bridges are changed from *State-$\alpha$* to *State-$\beta$* or to *State-$\gamma$*. In this regard, the DGCG model knows the mode shape changes that occurred in the states of source domains where the "bending" mode shape becomes "lateral-torsional", and the "lateral" mode becomes only "torsional". Yet, the mode shape changes are different in the target domains. For instance, the bending mode shape ($6^{th}$ mode) in *State-$\alpha$* of *Bridge#3* becomes "bending-longitudinal" or solely "longitudinal" in mode 7. In summary, if the types of new mode appearance/mode switch are different in the source domains than the target domains, this generally cause small values in CNF and occasionally a little lower MAC values.

**Observation 4**. The values in CNF in **Figure 20** (*Bridge#3*) are attributed to Observation 3, in which a different mode shape change occurs in mode five and mode 6 (**Table 3**) in the target domain than in the source domain (*Bridge#1*). However, CNF is zero in the comparison of *State-$\beta$* and *State-$\hat{\beta}$*, and *State-$\gamma$* and *State-$\hat{\gamma}$*, which is attributed to Observation 1. The low MAC values in mode 2 when comparing *State-$\alpha$* and *State-$\hat{\alpha}$* (for both cases) might also be attributed to Observation 1; however, the reasoning behind it is not fully clear. Additionally, MAC and average MMSC values are generally a little higher when comparing *State-$\beta$* and *State-$\hat{\beta}$*, and *State-$\gamma$* and *State-$\hat{\gamma}$*, than comparing *State-$\alpha$* and *State-$\hat{\alpha}$* (for both cases), which supports Observation 1.

**Observation 5**. The values in CNF in **Figure 21** (*Bridge#4*) are attributed to Observation 3 as different mode shape changes occur in modes 9 and 10 (**Table 3**) in the target domain than in the source domain (*Bridge#1*). The values in CNF are higher when comparing *State-$\alpha$* and *State-$\hat{\alpha}$* (for both cases) than the values in CNF for *State-$\beta$* and *State-$\hat{\beta}$*, and *State-$\gamma$* and *State-$\hat{\gamma}$*, which is attributed to Observation 1. Additionally, MAC and average MMSC values are generally a little higher when comparing *State-$\beta$* and *State-$\hat{\beta}$*, and *State-$\gamma$* and *State-$\hat{\gamma}$*, than comparing *State-$\alpha$* and *State-$\hat{\alpha}$* (for both cases), which also supports Observation 1.

**Observation 6**. *Bridge#2* is the most dissimilar to *Bridge#1*, which the DGCG model was trained on, as shown in **Figure 19**. The values in CNF are likely due to this very reason, supporting Observation 2. The values in CNF are slightly higher when comparing the *State-$\alpha$* and *State-$\hat{\alpha}$* (for both cases) than the values in CNF for *State-$\beta$* and *State-$\hat{\beta}$*, and *State-$\gamma$* and *State-$\hat{\gamma}$*, which is attributed to Observation 1. Similarly, the average MMSC values are higher when comparing *State-$\beta$* and *State-$\hat{\beta}$*, and *State-$\gamma$* and *State-$\hat{\gamma}$*, than the average MMSC values for *State-$\alpha$* and *State-$\hat{\alpha}$* (for both cases), which is also attributed to Observation 1. Additionally, *Bridge#2* is the stiffest among other bridges in which only two peaks in the singular values of spectral densities were identified. Thus, it was challenging to analyze *Bridge#2* due to the less identified modes.

**Observation 7**. The MMSC index is found to be a good index for SST for monitoring the training and testing of the model. As such, it shows great consistency with the natural frequencies, mode shapes, and MACs. For example, the average MMSC values are higher for the SSTs executed in Scenario I and Scenario III and lower in Scenario II, and Scenario IV, which was also the case concluded after checking the frequencies, mode shapes, and MACs (Observation 1). The MMSC values also show symmetry throughout the bridges from one half to the other half of the bridges, which makes sense as the bridges are geometrically and materially symmetrical, as well as the *State-$\beta$* and *State-$\gamma$*. Lastly, having an average MMSC value of 0.97 (97%) or higher (**Figure 20**) for comparing the states suggests that the modes of the bridge states are the same or extremely similar. This simplifies the SST evaluation process as it could optimize the effort spent for exhaustive modal identification procedures to check their natural frequencies, mode shapes and so on.

## 5. Summary and Conclusions

Implementation of SHM practices in every civil structure can be costly and impractical. Population-based SHM (PBSHM), a newly emerging research area, aims to increase the availability of physics and data-driven information on one set of civil structures based on the knowledge of other similar populations of civil structures. However, the studies presented in the literature do not adequately address the challenge of accessing the information (e.g., response data) of different structural states (conditions) of dissimilar civil structures.

On the other hand, ML methods have demonstrated favorable solutions to various challenges in the SHM field [76]. In a typical learning algorithm setting, the learner is trained on the *i.i.d.* assumption, meaning training and testing data are identically and independently distributed. However, in real-life problems, it is very likely that the learner faces a covariate and/or semantic shift in the target domains, which could deteriorate its performance. While acquiring data from all possible domains to train the learning algorithms is not practical, various research activity areas are suggested to deal with these shifts in domains (Out-Of-Distribution), such as Multi-Task Learning (MTL), Transfer-Learning (TL), Domain-Adaptation (DA), Test-Time Learning (TTL), Zero-Shot Learning (ZSL), Domain-Generalization (DG) so on. While TL, DA, and TTL leverage the information from the test (target) data in one way or another to expose the related knowledge to the model in training, DG has no access to the training (source) data and tests on the completely unseen domain, which makes it more challenging, providing more realistic and favorable results in real-world scenarios (**Table 1**).

The overall objective of the studies presented in the literature (explained in Section 1.1) is to adapt or transfer the knowledge from one set of the population of structure(s) to another by using DA or TL techniques. While the solutions offered in those studies were initially for homogenous populations, using DA or TL for heterogeneous populations has shown some success with promising results. The PBSHM is a relatively new research area, and a few studies exist, requiring additional studies to investigate to increase the availability of physics and/or data-driven knowledge on one set of civil structures based on the knowledge of another. In addition, while few studies are presented in the literature employing TL, DA, and ZSL approaches, using MTL, TTL, and particularly DG is not observed.

This study aims to estimate the response data of different civil structures based on the information obtained from a dissimilar structure. In this regard, "*Structural State Translation*" (SST) is introduced. SST is defined as *"Translating a state of one civil structure to another state after discovering and learning the domain-invariant representation in the source domains of a different civil structure"* (**Figure 2**). The SST framework is demonstrated on four dissimilar numeric bridge structure models under different scenarios

(**Figure 3**).

First, the Domain-Generalized Cycle-Generative (DGCG) model is trained to learn the domain-invariant representation in the acceleration datasets obtained from a numeric bridge structure, *Bridge#1*, that is in two different structural states: *State-α* and *State-β*. Then, the model is tested on three dissimilar numeric bridges to translate the bridges that are in *State-α* to *State-β, State-β* to *State-α*, *State-α* to *State-γ*, and *State-γ* to *State-α* where *State-α* is the pristine condition, *State-β* and *State-γ* are the removal of bottom chords from symmetric locations of the bridge, respectively. Since *State-β* and *State-γ* are structurally symmetrical, the translated *State-*$\hat{\beta}$ from *State-α* and the translated *State-*$\hat{\gamma}$ from *State-α* or vice versa are expected to be the same in terms of their structural parameters. Essentially, after the training of the DGCG model with the source domains of *State-α* and *State-β* of *Bridge#1*, the model is used to generalize and transfer its knowledge (the "domain-invariant representation") to other bridges, *Bridge#2*, *Bridge#3*, *Bridge#4* (unseen target data), which are structurally dissimilar.

The SST process for each translation scenario is evaluated using MMSC, a measure of similarity between signal pairs in the frequency domain, and modal identifiers of each real and translated state of *Bridge#2*, *Bridge#3*, and *Bridge#4*. The MMSC values revealed that the translated bridge states are extremely similar to the real states; as such, the lowest and highest average MMSC values obtained from the bridge states comparison are 91.2% and 97.1%, respectively. Additionally, significant similarity is observed between the natural frequencies and mode shapes of each real and translated bridge state. Hence, the highest and the lowest difference in natural frequencies among the modes of the bridge states are, respectively, 5.71% and 0%, while the highest and lowest MAC values are 0.998 and 0.870. Finally, several observations are made to clarify the SST process conducted for each translation scenario, which is explained in Section 4.

Several important conclusions are derived from this study and are listed below.

- SST procedure demonstrated quite well results in generating data for other dissimilar bridge structures.
- SSTs tested on three dissimilar bridges to generate their *State-β*/*State-γ* (unhealthy conditions) from *State-α* (healthy condition) was slightly better than generating *State-α* from *State-β*/*State-γ*.
- The DGCG model, which was trained on Bridge#1's states, performed better SST results on *Bridge#3* than on other bridges because *Bridge#1* and *Bridge#3* are most similar to each other regarding their stiffness/flexibility. It needs to be noted that all results indicate very high MMSC, MAC, and low differences in natural frequency comparisons.
- Generally, when the type of new mode appearance/mode switches between the source domains of

*State-α* and *State-β* are different from the ones in the target domains, those particular modes in the translated states are slightly off than the ones in real states.

- As modal identification procedures can be exhaustive for evaluating the SST results, the MMSC index is found to be an efficient index in this study as it gives quick intuition about the accuracy of SST results.
- Implementing SST on structurally symmetric structures can be very advantageous, as this study demonstrated that the outputs of SST for symmetric locations are identical, which could optimize the effort spent on data collection.
- Knowing the existence and the extent of domain-invariant representation in different domains remains an open question and an active research topic. Intuitively speaking, identical or similar structures should include domain invariancy; however, a domain relation, for example, between steel truss to prestressed concrete bridge structures or even bridge to dam structures, remains a big research question. Similarly, this statement is true for different damage types, severities, locations, or even the degree of linearities, as the structural nonlinearity in the structure may affect the data domains significantly.
- This study demonstrates that obtaining data from civil structures is possible while the structure is actually in different conditions or states. The SST procedure presented herein can potentially solve data scarcity and population-based applications in the SHM field.

**Glossary**

State: "State" and "condition" are interchangeably used to refer to the state/condition of the structure.

Domain: "Domain" refers to the class/category of the response data extracted from that particular state of the structure.

Translation: "Translation" refers to the domain translation process of the bridge states where the translated state is called the synthetic/translated/generated state.

State-α: "*State-α*" is the pristine condition of the bridge structures.

State-β: "*State-β*" is the condition where the bottom steel chords from both sides are removed from Section#11 of the bridge structures.

State-γ: "*State-γ*" is the condition where the bottom steel chords from both sides are removed from Section#12 of the bridge structures.

State-$\widehat{\alpha}/\widehat{\beta}/\widehat{\gamma}$: "*State*-$\hat{\alpha}/\hat{\beta}/\hat{\gamma}$" represent the translated states of that particular bridge structures.

**Acknowledgements**

This study was supported by the U.S. National Science Foundation (NSF) Division of Civil, Mechanical and Manufacturing Innovation (grant number 1463493), Transportation Research Board of The National Academies-IDEA Project 222, and National Aeronautics and Space Administration (NASA) Award No. 80NSSC20K0326 for the research activities and particularly for this paper.